\documentclass{article}

\PassOptionsToPackage{numbers,compress}{natbib}

\usepackage[main,final]{neurips_2026}

\usepackage[utf8]{inputenc}
\usepackage[T1]{fontenc}
\usepackage{hyperref}
\usepackage{url}
\usepackage{booktabs}
\usepackage{amsfonts}
\usepackage{nicefrac}
\usepackage{microtype}
\usepackage{xcolor}

\usepackage{graphicx}
\usepackage{mathtools,amssymb}
\usepackage{derivative}
\usepackage{algorithm}
\usepackage{algpseudocode}
\usepackage[capitalize]{cleveref}
\usepackage{multirow}
\usepackage{xspace}

\usepackage{pgfplots}\pgfplotsset{compat=1.18}
\usepgfplotslibrary{groupplots}%
\usepgfplotslibrary{fillbetween}%
\usetikzlibrary{matrix}        %
\usepackage{subcaption}
\definecolor{tabblue}{HTML}{1F77B4}
\definecolor{tabgreen}{HTML}{2CA02C}
\definecolor{tabred}{HTML}{D62728}
\pgfdeclareplotmark{hexagon}{%
  \pgfpathmoveto{\pgfpointpolar{0}{\pgfplotmarksize}}%
  \pgfpathlineto{\pgfpointpolar{60}{\pgfplotmarksize}}%
  \pgfpathlineto{\pgfpointpolar{120}{\pgfplotmarksize}}%
  \pgfpathlineto{\pgfpointpolar{180}{\pgfplotmarksize}}%
  \pgfpathlineto{\pgfpointpolar{240}{\pgfplotmarksize}}%
  \pgfpathlineto{\pgfpointpolar{300}{\pgfplotmarksize}}%
  \pgfpathclose\pgfusepathqfillstroke}

\providecolor{ColorGaU}{HTML}{42D4F4}
\providecolor{ColorIGMLE}{HTML}{4363D8}
\providecolor{ColorIGRobust}{HTML}{3CB44B}
\providecolor{ColorJeffreys}{HTML}{E6194B}
\providecolor{ColorMAGSAC}{HTML}{F58231}
\providecolor{ColorMSAC}{HTML}{FFE119}
\providecolor{ColorOracle}{HTML}{808080}
\providecolor{ColorRANSAC}{HTML}{FFB6C1}
\providecolor{ColorgauPerPairBest}{HTML}{42D4F4}
\providecolor{ColormagsacPerPairBest}{HTML}{F58231}
\providecolor{ColormsacPerPairBest}{HTML}{FFE119}
\providecolor{ColorransacPerPairBest}{HTML}{FFB6C1}

\pgfplotsset{ours_ransac/.style={color=ColorRANSAC, line width=1.0pt, solid, mark options={solid}}}
\pgfplotsset{ours_ransac_marker/.style={ours_ransac, mark=x, mark size=2pt, mark options={solid}}}
\pgfplotsset{ours_msac/.style={color=ColorMSAC, line width=1.0pt, solid, mark options={solid}}}
\pgfplotsset{ours_msac_marker/.style={ours_msac, mark=diamond*, mark size=2pt, mark options={solid}}}
\pgfplotsset{ours_gau/.style={color=ColorGaU, line width=1.0pt, solid, mark options={solid}}}
\pgfplotsset{ours_gau_marker/.style={ours_gau, mark=triangle*, mark size=2pt, mark options={solid}}}
\pgfplotsset{ours_magsac/.style={color=ColorMAGSAC, line width=1.0pt, solid, mark options={solid}}}
\pgfplotsset{ours_magsac_marker/.style={ours_magsac, mark=star, mark size=2pt, mark options={solid}}}
\pgfplotsset{ours_jeffreys/.style={color=ColorJeffreys, line width=1.0pt, solid, mark options={solid}}}
\pgfplotsset{ours_jeffreys_marker/.style={ours_jeffreys, mark=pentagon, mark size=2pt, mark options={solid}}}
\pgfplotsset{ours_ig_robust/.style={color=ColorIGRobust, line width=1.0pt, solid, mark options={solid}}}
\pgfplotsset{ours_ig_robust_marker/.style={ours_ig_robust, mark=o, mark size=2pt, mark options={solid}}}
\pgfplotsset{ours_ig_mle/.style={color=ColorIGMLE, line width=1.0pt, solid, mark options={solid}}}
\pgfplotsset{ours_ig_mle_marker/.style={ours_ig_mle, mark=square, mark size=2pt, mark options={solid}}}
\pgfplotsset{ours_oracle/.style={color=black, line width=1.0pt, solid}}

\newcommand{\BaselinesStdVtwoLo}{0.04}
\newcommand{\BaselinesStdVtwoHi}{0.38}
\newcommand{\StdGapVtwoLo}{2}
\newcommand{\StdGapVtwoHi}{29}
\newcommand{\ConvRatioPT}{128}
\newcommand{\ConvRatioETH}{8}

\newcommand{\EtoEOursMed}{1.57}   %
\newcommand{\EtoESRMed}{2.23}   %
\newcommand{\EtoEOursMAA}{0.69}   %
\newcommand{\EtoESRMAA}{0.64}   %
\newcommand{\EtoEOursMs}{50}   %
\newcommand{\EtoESRMs}{111}   %
\newcommand{\EtoETimePairs}{44}   %
\newcommand{\EtoETimeRepeats}{5}   %
\newcommand{\EtoEOursRange}{1.8\text{--}6.2}   %
\newcommand{\EtoESRRangeHi}{54}   %
\newcommand{\EtoESRInflMed}{38.27}   %
\newcommand{\SelStuMedCal}{2.32}   %
\newcommand{\SelSimfitMedCal}{2.36}   %
\newcommand{\SelStuMAACal}{0.623}   %
\newcommand{\SelSimfitMAACal}{0.622}   %
\newcommand{\SelStuMedInfl}{6.3}   %
\newcommand{\SelStuMAAInfl}{0.490}   %
\newcommand{\SelSimfitMedInfl}{34}   %
\newcommand{\SelSimfitMAAInfl}{0.134}   %
\newcommand{\PTValScenes}{10}   %
\newcommand{\PTTestScenes}{11}   %
\newcommand{\SelThreshMedInflLo}{37}   %
\newcommand{\SelThreshMedInflHi}{45}   %
\newcommand{\SelThreshMAAInfl}{0.049}   %

\DeclareMathOperator{\RSS}{RSS}

\makeatletter
\@for\@tmp:={A,B,C,D,E,F,G,H,I,J,K,L,M,N,O,P,Q,R,S,T,U,V,W,X,Y,Z,%
              a,b,c,d,e,g,h,i,j,k,l,m,n,o,p,q,r,s,t,u,v,w,x,y,z}\do{%
  \expandafter\edef\csname b\@tmp\endcsname{\noexpand\mathbf{\@tmp}}%
}
\@for\@tmp:={A,B,C,D,E,F,G,H,I,J,K,L,M,N,O,P,Q,R,S,T,U,V,W,X,Y,Z}\do{%
  \expandafter\edef\csname c\@tmp\endcsname{\noexpand\mathcal{\@tmp}}%
}
\@for\@tmp:={alpha,beta,gamma,delta,epsilon,varepsilon,zeta,theta,vartheta,%
              iota,kappa,lambda,mu,nu,xi,pi,varpi,rho,varrho,sigma,varsigma,tau,%
              upsilon,phi,varphi,chi,psi,omega,%
              Gamma,Delta,Theta,Lambda,Xi,Pi,Sigma,Upsilon,Phi,Psi,Omega}\do{%
  \expandafter\edef\csname b\@tmp\endcsname{%
    \noexpand\boldsymbol{\expandafter\noexpand\csname\@tmp\endcsname}%
  }%
}
\makeatother

\newcommand{\bzero}{\mathbf{0}}             %
\newcommand{\btSigma}{\widetilde{\bSigma}}  %
\newcommand{\NN}{\mathcal{N}}               %
\newcommand{\Reals}{\mathbb{R}}
\newcommand{\IG}{\mathrm{IG}}

\newcommand{\pin}{p_{\mathrm{in}}}
\newcommand{\pout}{p_{\mathrm{out}}}

\newcommand{\mname}[1]{\texorpdfstring{#1\xspace}{#1}}
\newcommand{\pp}{\nolinebreak[4]\hspace{-0.05em}+\nolinebreak[4]\hspace{-0.05em}+}
\newcommand{\ransac}{\mname{RANSAC}}
\newcommand{\msac}{\mname{MSAC}}
\newcommand{\mlesac}{\mname{MLESAC}}
\newcommand{\mapsac}{\mname{MAPSAC}}
\newcommand{\magsac}{\mname{MAGSAC}}
\newcommand{\magsacpp}{\texorpdfstring{MAGSAC\pp\xspace}{MAGSAC++}}
\newcommand{\gau}{\mname{GaU}}
\newcommand{\minpran}{\mname{MINPRAN}}
\newcommand{\orsa}{\mname{ORSA}}
\newcommand{\acransac}{\mname{AC-RANSAC}}
\newcommand{\recon}{\mname{RECON}}
\newcommand{\assc}{\mname{ASSC}}
\newcommand{\askc}{\mname{ASKC}}
\newcommand{\simfitpp}{\texorpdfstring{SIMFIT\pp\xspace}{SIMFIT++}}
\newcommand{\superansac}{\mname{SupeRANSAC}}
\newcommand{\loransac}{\mname{LO-RANSAC}}
\newcommand{\prosac}{\mname{PROSAC}}
\newcommand{\napsac}{\mname{NAPSAC}}
\newcommand{\degensac}{\mname{DEGENSAC}}
\newcommand{\qdegsac}{\mname{QDEGSAC}}
\newcommand{\gric}{\mname{GRIC}}

\newcommand{\poselib}{\mname{PoseLib}}
\newcommand{\igmle}{\mname{IG-MLE}}
\newcommand{\igrobust}{\mname{IG-Robust}}
\newcommand{\phototourism}{\mname{PhotoTourism}}
\newcommand{\ethd}{\mname{ETH3D}}
\newcommand{\heb}{\mname{HEB}}
\newcommand{\rootsift}{\mname{RootSIFT}}
\newcommand{\superpoint}{\mname{SuperPoint}}
\newcommand{\lightglue}{\mname{LightGlue}}
\newcommand{\gaussunif}{Gaussian--uniform\xspace}
\newcommand{\invgamma}{Inverse-Gamma\xspace}
\newcommand{\sortsweep}{sort-and-sweep\xspace}

\title{RANSAC Scoring Done Right}

\author{%
  James Pritts \quad Felix Seegräber \quad Kevin Köser \\
  Marine Data Science, Institut für Informatik \\
  Kiel University, Germany \\
  \texttt{\{jpr, fse, kk\}@informatik.uni-kiel.de} \\
}

\begin{document}

\maketitle

\begin{abstract}
	The most widely used \ransac variants score a candidate model by
	summing per-point scores that fall to zero beyond a threshold.
	These scores derive from the negative log-likelihood of a
	\gaussunif mixture. The threshold separates inliers, which
	follow the noise model, from outliers, and it is a function of the
	observation noise level, also known as the inlier scale. We remove
	this dependence on the threshold by reversing the usual order of
	\ransac inference: rather than estimating the scale and then
	scoring against it, we marginalize the scale in closed form for a
	fixed inlier partition and then optimize over partitions. The
	resulting score is maximized by a single sort followed by a linear
	sweep. A single closed-form expression spans informative and
	non-informative priors, so the same score adapts to datasets with
	extensive or limited ground-truth annotation. The proposed score
	has a single hyperparameter, the half-width of the uniform outlier
	distribution. We show that this half-width is far less sensitive
	to miscalibration than the inlier threshold and is easier to
	cross-validate. The proposed score is the first likelihood-based
	\ransac score in which the inlier scale is absent from the
	formula.\footnote{Project page, with code and the reproduction
		project: \url{https://ransacscoringdoneright.github.io}.}
\end{abstract}

\section{Introduction}
\label{sec:intro}

\ransac~\cite{Fischler1981} scores candidate geometric models by
evaluating how well they explain the observed data in the presence
of outliers.  Every widely used scoring function, whether inlier
counting (\ransac), truncated squared error (\msac~\cite{Torr1998}),
or the \gaussunif marginal likelihood
(\mlesac~\cite{Torr2000}), depends on a hyperparameter that
implicitly encodes the inlier scale.  The user
must either supply this quantity or estimate it from
contaminated data, neither of which is straightforward in
practice.

Shekhovtsov~\cite{Shekhovtsov2025} provides a unifying analysis of
these functions through the lens of a single probabilistic
model: the \gaussunif mixture, in which each inlier
residual follows a Gaussian density and each outlier is uniform.
Within this framework, \msac is the \emph{profile} likelihood
(maximizing over latent inlier labels) and \mlesac is the
\emph{marginal} likelihood (summing over labels), the score
Shekhovtsov calls \gau.  Both are
parameterized by a decision threshold $\tau = q\,\sigma$, a fixed
quantile~$q$ of the inlier scale~$\sigma$, so tuning the threshold is equivalent to
tuning the inlier scale in different units.
We call scores of this form \emph{threshold-based}.

We propose an \emph{inlier-scale-free} \ransac score, the
\emph{profile-marginal score}, by reversing the
usual order of elimination.  Threshold-based scores sum or maximize
over the latent inlier/outlier labels first, retaining the inlier scale
through the threshold.  We instead integrate out the inlier
scale~$\sigma$ for each fixed partition and then maximize over
partitions.  To our knowledge, this is the first likelihood-based
\ransac score to eliminate~$\sigma$ in this order.  A conjugate
\invgamma prior on~$\sigma^2$ gives a closed form that also covers
the Jeffreys limit.  When annotated validation data are available, we
estimate the prior parameters by empirical Bayes.  Otherwise, we use
the Jeffreys limit with the same scoring algorithm.

For a fixed prior, the score depends on the inlier residual sum of
squares, the inlier count, and the half-width~$a$ of the outlier
domain (\Cref{sec:scoring}).  We maximize it by sorting the
point-to-model distances; the best inlier set is always a prefix of
the sorted list, so one sweep over the prefixes finds it.  This takes
$O(N)$ time with the radix sort used in our implementation and
$O(N\log N)$ with a comparison sort.  The half-width~$a$ parameterizes
the outlier model and is not tied to any quantile of the inlier scale.
When an image pair contains enough inliers for the data to dominate
the prior, we find that the score is significantly less sensitive
to~$a$ than threshold-based scores are to their threshold.  The score
is thus inlier-scale-free, and \Cref{sec:supp-torr} compares our score with Torr's
Bayesian framework and \gric~\cite{Torr2002}.

\magsac~\cite{Barath2019} and \magsacpp~\cite{Barath2020} were
introduced specifically to remove the inlier-scale dependence by
``marginalizing'' the inlier scale under the \gaussunif
assumption.  Shekhovtsov~\cite{Shekhovtsov2025} shows that the
derivation's errors cancel by numerical coincidence, leaving a score
numerically equivalent to that of \mlesac with a fixed inlier
scale. The \magsacpp hyperparameter is defined as the upper bound on
the inlier scale, but it is in fact a proxy for the mixture's inlier
scale (\Cref{sec:supp-torr}).  Moreover, when the threshold is
properly tuned, \emph{all} of these scoring functions (\msac, \gau,
\magsacpp) perform nearly identically for model selection. Only inlier
counting (standard \ransac) is measurably worse.  The bottleneck is not
the choice of scoring function but the dependence on a well-chosen
inlier-scale quantile.

Other work avoids a hand-set threshold.  Adaptive-scale estimators
(\assc~\cite{Wang2004}, \askc~\cite{Wang2010}) and
\simfitpp~\cite{Edstedt2025} estimate the scale from the contaminated
distances and then score against it. This is the estimate-then-score
order we reverse.  \recon~\cite{Raguram2011} instead seeks consistency
in how different models rank the points by distance.  A-contrario
methods (\minpran~\cite{Stewart1995},
\orsa/\acransac~\cite{Moisan2004,Moisan2012}) are closest in spirit:
they also sweep distance-sorted prefixes and require only an outlier
domain, but they rank inlier sets by a significance test against a
uniform background model rather than by a likelihood.  Our score is
the marginal likelihood of the same \gaussunif model that the
threshold-based scores share.

\section{Probabilistic Model for Robust Fitting}
\label{sec:partition}
Consider $N$ observations $\bx_i \in \Reals^d$, $i = 1,\ldots,N$, and
a candidate geometric model~$\btheta \in \Reals^{n_\theta}$.  The
noise-free point~$\bar\bx_i$ of each inlier lies on the model manifold
$\cM_{\btheta} = \{\bx \in \Reals^d : \bg(\bx,\btheta) = \bzero\}$,
whose codimension $d_g = d - \dim\cM_{\btheta}$ is the number of
directions normal to it.  Noise along~$\cM_{\btheta}$ leaves~$\bg$
unchanged to first order, so only the $d_g$ normal directions carry
evidence.  The codimension therefore fixes the dimension of the
residual vector, and each observation imposes $d_g$ independent scalar
constraints on~$\btheta$, one per normal direction.  A minimal sample
has $m = \lceil n_\theta/d_g \rceil$ observations, the fewest whose
constraints determine~$\btheta$ up to finitely many solutions.  A
minimal solver computes these solutions in the hypothesis step that
proposes models for scoring; \Cref{tab:model-dims} lists $(n_\theta,
	d_g, m)$ for common geometric models.

An inlier is a noisy observation of a point on the model, $\bx_i =
	\bar\bx_i + \bvarepsilon_i$ with $\bg(\bar\bx_i,\btheta) = \bzero$ and
$\bvarepsilon_i \sim \NN(\bzero, \sigma^2\btSigma_{\bx_i})$, where the
shape $\btSigma_{\bx_i}$ is known and the inlier scale~$\sigma$ is
not.  To first order $\bg_i \coloneqq \bg(\bx_i,\btheta) \approx
	\bJ_i\bvarepsilon_i$, with $\bJ_i$ the Jacobian of~$\bg$ with respect
to~$\bx$. Whitening by $\bL_i\bL_i^\top =
	\bJ_i\btSigma_{\bx_i}\bJ_i^\top$ gives the residual
\begin{equation}
	\br_i \coloneqq \bL_i^{-1}\bg_i \;\sim\; \NN(\bzero,
	\sigma^2\bI_{d_g}),
	\label{eq:residual}
\end{equation}
whose squared norm
$\bg_i^\top(\bJ_i\btSigma_{\bx_i}\bJ_i^\top)^{-1}\bg_i$ is the general
form of the squared Sampson
distance~\cite{Sampson1982,Hartley2003,Chojnacki2000}. Equivalently,
it is the squared Mahalanobis distance, with respect
to~$\btSigma_{\bx_i}$, from~$\bx_i$ to the linearization
of~$\cM_{\btheta}$ at~$\bx_i$.  Its relation to the geometric distance
to~$\cM_{\btheta}$ is discussed in~\cite{Harker2006,Rydell2024}. Our
model requires only that the distribution of the residual be isotropic
normal.

We adopt the standard \gaussunif
mixture~\cite{Torr2000,Shekhovtsov2025}: a latent label $z_i \in
	\{0,1\}$ marks observation~$i$ as inlier ($z_i{=}1$) or outlier
($z_i{=}0$) with prior $\Pr(z_i{=}1) = \gamma$, and the
residual~$\br_i$ has conditional density
\begin{equation}
	p(\br_i \mid z_i, \btheta, \sigma) = \begin{cases}
		\pin(\br_i;\sigma) \coloneqq (2\pi\sigma^2)^{-d_g/2}\,
		\exp\!\bigl({-}\|\br_i\|^2/2\sigma^2\bigr) & z_i = 1, \\[3pt]
		\pout \coloneqq (2a)^{-d_g}                & z_i = 0,
	\end{cases}
	\label{eq:component-densities}
\end{equation}
a Gaussian inlier density and a uniform outlier density on
$[-a,a]^{d_g}$; the half-width~$a$ has the same units as~$\sigma$
(e.g.\ pixels for a Sampson distance).  We normalize coordinates so
that the image half-diagonal is one, which fixes the outlier domain,
and hence~$a$, independently of~$\sigma$.  By independence, the joint
density of residuals and labels $\bz = (z_1,\ldots,z_N)$ factorizes as
\begin{equation}
	p(\br, \bz \mid \btheta, \sigma) = \prod_{i=1}^{N} \bigl[\gamma\,
		\pin(\br_i;\sigma)\bigr]^{z_i}\, \bigl[(1{-}\gamma)\,
		\pout\bigr]^{1-z_i}.
	\label{eq:joint}
\end{equation}
Maximizing~\eqref{eq:joint} over labels classifies $z_i^*=1$ iff
$\gamma\,\pin(\br_i;\sigma) > (1{-}\gamma)\,\pout$; collecting
positive labels defines the inlier set $\cI \coloneqq \{i : z_i^* =
	1\}$ with complement $\cO = \{1,\ldots,N\}\setminus\cI$, and $n_I =
	|\cI|$, $n_O = |\cO|$.  This per-point rule is the profile-likelihood
labeling at fixed~$\sigma$; in \Cref{sec:scoring} we instead treat the
partition~$\cI$ as a free variable to optimize.  Given a partition,
the likelihood factorizes as
\begin{equation}
	p(\br \mid \btheta, \cI, \sigma) = \prod_{i \in \cI}
	\frac{1}{(2\pi\sigma^2)^{d_g/2}}\,
	\exp\!\Bigl({-}\frac{\|\br_i\|^2}{2\sigma^2}\Bigr) \;\cdot\;
	\frac{1}{(2a)^{n_O d_g}}\,.
	\label{eq:joint-partition}
\end{equation}

\section{Profile-Marginal Likelihood Scoring}
\label{sec:scoring}

Besides the model~$\btheta$, the joint~\eqref{eq:joint} has two
nuisance quantities, the labels~$\bz$ and the scale~$\sigma$.
Scoring a candidate requires eliminating both.  Standard
approaches eliminate the labels first---by summation (\mlesac's
marginal likelihood~\cite{Torr2000}) or maximization (\msac's
profile likelihood~\cite{Torr1998})---and the resulting scores
still depend on~$\sigma$~\cite{Shekhovtsov2025}.

\subsection{Reversing the Elimination Order}
\label{sec:reverse-order}

We take a different path.  Rewrite the
joint~\eqref{eq:joint} by grouping labels into a partition
$\cI = \{i : z_i = 1\}$:
\begin{equation}
	p(\br, \bz \mid \btheta, \sigma)
	= \gamma^{n_I}(1{-}\gamma)^{n_O}
	\;\cdot\; p(\br \mid \btheta, \cI, \sigma),
	\label{eq:joint-partition-decomp}
\end{equation}
where $p(\br \mid \btheta, \cI, \sigma)$ is the partition-conditioned
likelihood~\eqref{eq:joint-partition} and the label prior
$\gamma^{n_I}(1{-}\gamma)^{n_O}$ depends only on~$n_I = |\cI|$.  For
general~$\gamma$ the label prior adds $n_I\log\!\tfrac{\gamma}{1-\gamma}$
(plus a constant) to the log-score, biasing the inlier/outlier
boundary.  We adopt the non-informative choice $\gamma = \tfrac{1}{2}$,
for which the label prior is the constant $2^{-N}$ across all
partitions and drops out of the optimization.\footnote{One could instead
	marginalize~$\gamma$ under a conjugate Beta prior, exactly as we
	marginalize~$\sigma$: $\int_0^1
		\gamma^{n_I}(1{-}\gamma)^{n_O}\,\pi(\gamma)\,d\gamma$ is closed-form
	and, depending only on the count~$n_I$, preserves the \sortsweep
	optimality and its cost.  We fix $\gamma=\tfrac12$ instead so that the
	label prior is partition-independent and the consensus decision is
	driven solely by fit quality and outlier volume rather than a
	separately estimated inlier rate, which itself depends on the
	partition.}

Rather than eliminating labels first (which entangles~$\sigma$ with
the per-point sums or maxima), we \emph{reverse the order}: for a
\emph{fixed} partition~$\cI$ and a fixed candidate~$\btheta$, a plug-in
for the minimal-sample solution as in all \ransac scores
(\Cref{sec:supp-dof}), we marginalize~$\sigma$ under a conjugate
\invgamma prior $\pi(\sigma^2) = \IG(\alpha_0, \beta_0)$, defining the
\emph{scale-marginalized log-likelihood}
\begin{equation}
	S(\btheta, \cI)
	\;\coloneqq\; \log \int_0^\infty
	p(\br \mid \btheta, \cI, \sigma)\;
	\pi(\sigma^2)\, d\sigma^2,
	\label{eq:S-def}
\end{equation}
and then optimize over partitions, giving the model score
\begin{equation}
	S^*(\btheta) \coloneqq \max_{\cI} \; S(\btheta, \cI).
	\label{eq:profile-I}
\end{equation}
We call~$S^*$ the \emph{profile-marginal score}: profile over
labels~$\bz$ (equivalently partitions~$\cI$), marginal
over~$\sigma$.  A fully Bayesian score would marginalize
\emph{both}, but $\log \sum_{\cI} \exp S(\btheta,\cI)$ has $2^N$
terms; the hybrid avoids this sum while still eliminating~$\sigma$
analytically, because for a fixed partition the
$\sigma$-integral~\eqref{eq:S-def} has a closed form
(\Cref{sec:score}).

\subsection{The Profile-Marginal Score in Closed Form}
\label{sec:score}

The \invgamma is the conjugate prior for the variance of a zero-mean
Gaussian, so~\eqref{eq:S-def} has a closed form for any
$\alpha_0, \beta_0 > 0$; we keep $(\alpha_0, \beta_0)$ general, so the
score holds for every prior in the family.  Taking logarithms yields
the closed form of~$S$, from which the profile-marginal
score~\eqref{eq:profile-I} follows by maximizing over~$\cI$
(\Cref{sec:supp-score-deriv} gives the full derivation):
\begin{equation}
	\boxed{
		S(\btheta, \cI)
		= \log\Gamma\!\Bigl(\frac{\nu}{2} + \alpha_0\Bigr)
		- \Bigl(\frac{\nu}{2} + \alpha_0\Bigr)
		\log\!\Bigl(\frac{\RSS_\cI}{2} + \beta_0\Bigr)
		- \frac{\nu}{2}\log(2\pi)
		- n_O\, d_g\,\log(2a),
	}
	\label{eq:score}
\end{equation}
where $\RSS_\cI = \sum_{i\in\cI} \|\br_i\|^2$ is the inlier residual
sum of squares and $\nu = n_I\,d_g - n_\theta$ its degrees of freedom:
the $n_I\,d_g$ Gaussian residual coordinates of the inliers less the
$n_\theta$ consumed by fitting~$\btheta$ (\Cref{sec:supp-dof} derives
this correction).

\paragraph{Pseudo-observation reading.}\label{par:pseudo-obs}
The integrand of~\eqref{eq:S-def} is proportional to an \invgamma
density in~$\sigma^2$, so the posterior given the inlier residuals is
\begin{equation}
	\sigma^2 \mid \cI \;\sim\;
	\IG\!\Bigl(
	\tfrac{2\alpha_0 + \nu}{2},\;
	\tfrac{2\beta_0 + \RSS_\cI}{2}
	\Bigr).
	\label{eq:ig-posterior}
\end{equation}
Each inlier~$i$ outside the minimal sample contributes $d_g$ residual
coordinates to~$\nu$ and $\|\br_i\|^2$ to~$\RSS_\cI$ (so
$\nu = (n_I - m)\,d_g$ for an exact minimal solver, \Cref{sec:supp-dof}).  Without prior
information (the Jeffreys limit) the posterior is
$\IG(\nu/2, \RSS_\cI/2)$; comparing with~\eqref{eq:ig-posterior}, the prior only replaces the
coordinate count~$\nu$ by $\nu + 2\alpha_0$ and the sum of squares
$\RSS_\cI$ by $\RSS_\cI + 2\beta_0$.  It thus enters the posterior
exactly as $2\alpha_0$ additional residual coordinates would, the
\emph{pseudo-observations}, with sum of squares
(\emph{pseudo-RSS}) $2\beta_0$.  Their mean square
$s_0^2 = 2\beta_0/2\alpha_0 = \beta_0/\alpha_0$ is the prior's guess
of~$\sigma^2$, just as $\RSS_\cI/\nu$ is the data's.  Whether fitted by empirical
Bayes (\Cref{sec:supp-prior-estimation} gives two estimators) or chosen
by hand, the prior thus specifies both $s_0^2$ and the
number $2\alpha_0$ of coordinates supporting it.  As a point estimate of the
inlier variance we take the mean square of the real and
pseudo-observations together, which reduces to the classical
$\RSS_\cI/\nu$ in the Jeffreys limit (and equals
$1/\mathbb{E}[\sigma^{-2} \mid \cI]$, \Cref{sec:supp-score-deriv});
for $\nu > 0$,
\begin{equation*}
	\hat s^2 \coloneqq \frac{2\beta_0 + \RSS_\cI}{2\alpha_0 + \nu}
	= \frac{2\alpha_0}{2\alpha_0 + \nu}\,s_0^2
	+ \frac{\nu}{2\alpha_0 + \nu}\,\frac{\RSS_\cI}{\nu}.
\end{equation*}
Marginally, an inlier residual coordinate is then Student-$t$ with
$2\alpha_0 + \nu$ degrees of freedom and scale~$\hat s$.  The prior
mean square dominates when $\nu \ll 2\alpha_0$ and the observed one
when $\nu \gg 2\alpha_0$, which is how the prior regularizes the variance
estimate in the data-scarce regime of \Cref{sec:experiments}.

\paragraph{The four terms of the score.}
All four terms of~\eqref{eq:score} depend on the partition~$\cI$:
\begin{enumerate}
	\item \emph{Inlier count.} $\log\Gamma(\nu/2 + \alpha_0)$ grows
	      super-linearly for large~$\nu$, favoring larger consensus sets.
	\item \emph{Fit quality.}
	      $-(\nu/2 + \alpha_0)\log(\RSS_\cI/2 + \beta_0)$ favors smaller
	      residual sums of squares at fixed~$\nu$.  A positive pseudo-RSS
	      $2\beta_0$ keeps the term finite as $\RSS_\cI \to 0$.
	\item \emph{Gaussian normalizer.} $-(\nu/2)\log(2\pi)$ supplies the
	      normalization constant for the $\nu$ inlier coordinates.
	\item \emph{Outlier penalty.} $-n_O\,d_g\log(2a)$ assigns each outlier
	      the uniform-model cost $d_g\log(2a)$.
\end{enumerate}

\paragraph{Non-informative limit (Jeffreys prior).}
The limit $(\alpha_0, \beta_0) \to (0,0)$ gives the improper Jeffreys
prior $\pi(\sigma^2) \propto 1/\sigma^2$~\cite{Jeffreys1946}, which
introduces no scale assumption. Its posterior is proper whenever $\nu
	> 0$, so we restrict the maximization in~\eqref{eq:profile-I} to
partitions with $\nu > 0$, i.e.\ $n_I > n_\theta/d_g$. For example,
with an exact minimal solver, at least one inlier beyond the minimal
sample is required. The score~\eqref{eq:score} then simplifies to
\begin{equation}
	S_J(\btheta, \cI) = \log\Gamma\!\Bigl(\frac{\nu}{2}\Bigr) -
	\frac{\nu}{2}\,\log(\pi\,\RSS_\cI) - n_O\, d_g\,\log(2a),
	\label{eq:score-jeffreys}
\end{equation}
where the combination $(\nu/2)\log(2\pi) + (\nu/2)\log(\RSS_\cI/2) =
	(\nu/2)\log(\pi\,\RSS_\cI)$ has been merged.  Every term is data-only,
without any pseudo-observation contribution from the prior, so $S_J$
is the natural baseline against which an informative prior adds value
only when $\nu$ is small, i.e.\ when there are too few inlier
residuals to constrain~$\sigma$ (the weighted average above).

\paragraph{Implicit threshold.}
Although the score eliminates~$\sigma$, the boundary of the optimal
partition~$\cI^*$ defines an \emph{implicit} threshold that, in the
large-sample limit, satisfies the MAP condition $\pin(\br;\hat s) =
	\pout$ at $\|\br\| = \tau^*$ (the inlier density depends on~$\br$ only
through $\|\br\|$), with $\hat s$ the variance estimate above evaluated at~$\cI^*$
($\hat s^2 = \RSS_{\cI^*}/\nu$ in the Jeffreys limit):
\begin{equation}
	\tau^* = \hat s\,\sqrt{2\,d_g\log\bigl(a\sqrt{2/\pi}\,/\,\hat s\bigr)},
	\qquad
	\frac{\partial\log\tau^*}{\partial\log a}
	= \frac{1}{2\log\bigl(a\sqrt{2/\pi}\,/\,\hat s\bigr)}.
	\label{eq:implicit-threshold}
\end{equation}
Unlike the baselines' one global~$\tau$, the proposed profile-marginal
score adapts per hypothesis through~$\hat s$
(\Cref{sec:implicit-threshold} derives it).  It also
explains the robustness to miscalibration: a baseline's boundary is
proportional to its hyperparameter, whereas $\tau^*$ depends on~$a$
only through a square-rooted logarithm:
by~\eqref{eq:implicit-threshold}, for any~$d_g$, a 1\% change in~$a$
changes~$\tau^*$ by only 0.1--0.25\%, and even overestimating~$a$ tenfold widens~$\tau^*$ by
only 24--45\%.  The threshold is the MAP threshold of the \gaussunif mixture at
$\gamma = \tfrac12$,
and for $a/\hat s \in [10, 100]$ it gives
$\tau^* \approx (2\text{--}3)\sqrt{d_g}\,\hat s$, recovering the familiar
$\chi^2$-quantile rules of thumb ($1.96\sigma$--$3\sigma$ for $d_g = 1$,
$2.45\sigma$--$3.4\sigma$ for $d_g = 2$~\cite{Hartley2003}) as a
Bayes decision boundary, which balances rejected inliers against
accepted outliers, rather than one that fixes only the rate of rejected
inliers: the
boundary sits where the Gaussian and the outlier density are equally
likely, and tightens automatically as outliers become denser.

\subsection{The Sort-and-Sweep Algorithm}

For a fixed~$\btheta$, the distances $\delta_i = \|\br_i\|$
depend only on~$\btheta$, not on the partition.  Sorting them
$\delta_{(1)} \leq \cdots \leq \delta_{(N)}$, the optimal inlier set is
always a prefix $\cI_k = \{(1),\ldots,(k)\}$ of some length~$k$,
written $\cI^* = \cI_{k^*}$ at the optimum: swapping any
included point for an excluded one with a smaller distance
lowers~$\RSS_\cI$ at fixed~$n_I$ and strictly raises the score.
Since $\RSS_{\cI_k} = \sum_{j=1}^{k} \delta_{(j)}^2$ is then a prefix sum, a
running accumulator evaluates the full score in one $O(N)$ pass after
the sort, starting at $k = \lfloor n_\theta/d_g \rfloor + 1$, the smallest
length with
$\nu = k\,d_g - n_\theta > 0$ (\Cref{sec:score}); the Jeffreys limit ($\alpha_0{=}\beta_0{=}0$) is the
same algorithm without the pseudo-observations.
\Cref{alg:score} summarizes the
procedure.  Within \ransac, the
sweep is applied to each candidate model generated from a
minimal sample~\cite{Fischler1981}; the model with the highest
score~$S^*$ is selected.  The sort sets the cost: $O(N\log N)$ with
a comparison sort, and $O(N)$ with a radix sort on fixed-precision
keys (we sort 32-bit floats, the precision of the residuals), the
same order as \msac at 9--20\,$\mu$s more per hypothesis
(\Cref{tab:runtime-hyp}).

\begin{algorithm}[h]
	\caption{Sort-and-sweep computation of the
		profile-marginal score~\eqref{eq:profile-I}, using~\eqref{eq:score}.
		The algorithm returns the best-scoring prefix
		$\cI^* = \{(1),\ldots,(k^*)\}$ with one sort and an $O(N)$ sweep.}
	\label{alg:score}
	\begin{algorithmic}[1]
		\Require distances
		$\delta_1,\ldots,\delta_N$;
		codimension~$d_g$;
		model parameters~$n_\theta$;
		outlier half-width $a$;
		prior parameters $(\alpha_0, \beta_0)$
		\Ensure best score $S^*$; optimal inlier count $k^*$
		\State Sort so that
		$\delta_{(1)} \leq \delta_{(2)} \leq \cdots \leq \delta_{(N)}$
		\State $k_{\min} \gets \min\!\bigl(\lfloor n_\theta/d_g \rfloor,\, N\bigr)$
		\Comment{largest prefix with $\nu \leq 0$}
		\State $\RSS \gets \sum_{j=1}^{k_{\min}} \delta_{(j)}^2$
		\Comment{seed the zero-DOF prefix}
		\State $S^* \gets -\infty$, \quad $k^* \gets 0$
		\For{$k = k_{\min}+1, \ldots, N$}
		\State $\RSS \gets \RSS + \delta_{(k)}^2$
		\State $\nu \gets k\,d_g - n_\theta$
		\Comment{positive DOF guaranteed}
		\State $S \gets \log\Gamma\!\bigl(\nu/2 + \alpha_0\bigr)
			- \bigl(\tfrac{\nu}{2}+\alpha_0\bigr)\log\!\bigl(\tfrac{\RSS}{2}+\beta_0\bigr)
			- \tfrac{\nu}{2}\log(2\pi)
			- (N{-}k)\,d_g\,\log(2a)$
		\If{$S > S^*$}
		\State $S^* \gets S$, \quad $k^* \gets k$
		\EndIf
		\EndFor
	\end{algorithmic}
\end{algorithm}

\section{Experiments}
\label{sec:experiments}

State-of-the-art \ransac pipelines achieve performance through
the combined effect of multiple components: guided sampling
(\prosac~\cite{ChumMatas2005}, \napsac~\cite{Myatt2002napsac}),
degeneracy testing
(\degensac~\cite{Chum2005degensac}, \qdegsac~\cite{FrahmPollefeys2006qdegsac}),
local optimization
(\loransac~\cite{Chum2003,Lebeda2012}), and
varied termination criteria~\cite{Schoenberger2025stopping}.
Many of these components are domain-specific (e.g., guided
sampling for two-view matching~\cite{Barath2019pnapsac}) and do
not transfer to other model classes.  Proper
ablation of the score from the rest is rarely performed, so a
method's apparent advantage may stem from other components
rather than the scoring strategy~\cite{Shekhovtsov2025}.  We follow~\cite{Shekhovtsov2025} in isolating the score, with
some adjustments called out in the following paragraphs.  The
takeaway is that every method evaluates the same per-pair pool
of minimal-sample hypotheses, differing only in its scoring
function and its single hyperparameter.

We evaluate each scoring method along two
complementary axes.
The \emph{Profile} experiment (\Cref{sec:profile}) sweeps each
method's hyperparameter and measures \emph{robustness to
	hyperparameter value}, the flatness of the error under
perturbation of the optimal parameter on the validation set.
The \emph{Sensitivity} experiment (\Cref{sec:sensitivity-expt})
varies only the validation-set size and measures the smallest
validation set sufficient to reach near-optimal test
performance.

\subsection{Evaluation Protocol}
\label{sec:eval-protocol}

For each image pair we sample $1\,000$ two-view geometric
relations (homography, essential, or fundamental matrix,
depending on the dataset) from pre-extracted feature
correspondences.  Each method scores this shared model pool;
the highest-scored model is evaluated against the ground-truth
transformation by \emph{orientation error}, which is the
$\mathrm{SO}(3)$ angular distance between the recovered and
oracle rotations.  The rotation is recovered per model class:
directly from the essential matrix, and via the known
calibration for the fundamental matrix ($\bF\!\to\!\bE$) and the
homography (by decomposing $\bH$).  Both Profile and Sensitivity use this same
metric.  Datasets group image pairs by scene.  The Sensitivity
experiment (\Cref{sec:sensitivity-expt}) draws its
validation/test split hierarchically by scene on \heb---resampling
sites, then pairs within each, to test cross-scene
generalization---and by a flat pair-level split on \phototourism
and \ethd; the Profile bands (\Cref{sec:profile}) are pair-level
on all three (\Cref{tab:datasets}).

Shekhovtsov~\cite{Shekhovtsov2025} showed that the four
threshold-based scoring functions we compare---RANSAC~\cite{Fischler1981},
\msac~\cite{Torr1998}, \gau~\cite{Torr2000}, and
\magsacpp~\cite{Barath2020}---can each be reparameterized with a
single hyperparameter.  Our three proposed marginal-likelihood variants (Jeffreys,
\igrobust, \igmle) likewise share a single hyperparameter, the
outlier half-width from~\eqref{eq:score}.  We write~$\kappa$ for
each method's single hyperparameter generically (the half-width
$a$ for our profile-marginal scores, the threshold quantile for the
baselines), normalized to its per-dataset optimum~$\kappa^\star$
in all plots.  However, the physical interpretation of the role of the
hyperparameter in each family is different, and we should not
expect them to share the same numeric range nor do we constrain
them to.

Shekhovtsov~\cite{Shekhovtsov2025} hand-sets per-method search
intervals and shows that even a generous 2+~decade interval
based on a reasonable guess can disadvantage some methods,
since the interval acts as a weak hyperprior on the
hyperparameter.  We instead derive every method's interval
from data: we draw a single validation set per dataset and
evaluate each method's validation-set orientation error on a coarse
log-spaced grid; the minimum-error grid point becomes the
anchor~$\kappa^\star$ (if $\kappa^\star$ lies on a boundary, the search extends
by two more decades).  The two-decade window
$[\kappa^\star/10,\,\kappa^\star\cdot 10]$ is the per-method, per-dataset
interval used for hyperparameter selection in both the
\emph{Profile} (\Cref{sec:profile}) and \emph{Sensitivity}
(\Cref{sec:sensitivity-expt}) experiments.  To parallelize
scoring we sample 100 points per decade uniformly in log-space
within this window (200 points total); hyperparameter
selection in both experiments is done over this fixed grid.
All seven methods use the same resolution and the same
two-decade width.

Our empirical-Bayes variants (\igmle and \igrobust) need
ground-truth annotations---the relative pose and a set of
ground-truth point matches per pair---during the validation
phase.  Both fit the \invgamma prior on the noise
variance~$\sigma^2$ from per-pair noise-variance
estimates~$\hat\sigma^2_p$ on the validation
set~(\Cref{sec:supp-prior-estimation}).  \emph{\igmle} fits the
prior by type-II maximum likelihood: since
$\sigma^2\!\sim\!\IG(\alpha,\beta) \iff
	1/\sigma^2\!\sim\!\mathrm{Gamma}(\alpha,1/\beta)$, it fits a
Gamma to $\{1/\hat\sigma^2_p\}$ and reads off
$(\alpha_0,\beta_0)$, capturing both the location and the
dispersion of the $\hat\sigma^2_p$ but remaining sensitive to
outlier per-pair estimates.  \emph{\igrobust} instead fixes
$\alpha_0 = 1$ (two pseudo-observations) and sets
$\beta_0 = \alpha_0\,\mathrm{median}(\hat\sigma^2_p)$, trading
dispersion information for robustness to the heavy
$\hat\sigma^2_p$ tails that arise when match-set cardinality
approaches the model's degrees of freedom (demonstrated on
\heb~\cite{Barath2023} in \Cref{sec:sensitivity-expt}).  Each
$\hat\sigma^2_p$ is computed from residuals on that pair's GT
matches under the GT pose; how those annotations are obtained
is dataset-specific and detailed in \Cref{sec:supp-datasets}.

\subsection{Datasets}
\label{sec:datasets}

We evaluate on three datasets, one per two-view geometric
model class, with model proposals generated from both a
hand-crafted and a state-of-the-art learned feature detector
(nearly $70\,000$ pairs in total), spanning three axes of
variation: dataset, geometric problem, and feature pipeline.
\emph{\phototourism} (PT)~\cite{Jin2021,RansacTutorial2020}
supplies 21 landmark scenes (44\,211 pairs) with
\rootsift~\cite{Arandjelovic2012} features for essential-matrix
($\bE$) estimation; \emph{\ethd}~\cite{Schoeps2017} supplies 13
sub-scenes (3\,039 pairs) with learned
\superpoint\,+\,\lightglue~\cite{DeTone2018,Lindenberger2023}
features for fundamental-matrix ($\bF$) estimation; and the
\emph{Homography Estimation Benchmark} (\heb)~\cite{Barath2023}
supplies 10 architectural sites (21\,999 pairs) with
SIFT~\cite{Lowe2004} features for homography ($\bH$)
estimation.  The empirical-Bayes variants (\igmle, \igrobust)
additionally consume per-pair ground-truth matches during
validation (\Cref{sec:eval-protocol}).  Full dataset
provenance, annotation and feature-pipeline details, and the
\heb plane-assignment caveat that surfaces in some \heb panels
below are given in \Cref{sec:supp-datasets}
(\Cref{tab:datasets}).

\subsection{Profile}
\label{sec:profile}

The Profile experiment plots each method's median rotation
error across all pairs in the dataset as a function of its
hyperparameter, swept exhaustively over the two-decade
interval described in \Cref{sec:eval-protocol}.  It is not a
tuning protocol (\Cref{sec:sensitivity-expt} gives a practical one):
each method's minimum is its achievable lower bound, and the rate at
which error climbs off-optimum is a direct readout of sensitivity to
miscalibration.  No model is selected, so evaluating on the full pool
involves no validation--test leakage.

\begin{figure}[t]
	\centering
	\input{figures/profile_styles}%
	\pgfplotsset{panel_axis/.append style={
				label style={font=\scriptsize},
				title style={font=\scriptsize},
				tick label style={font=\tiny}}}%
	\renewcommand{\PanelXLabel}{$\kappa / \kappa^\star$}%
	\begin{tikzpicture}
		\begin{groupplot}[panel_axis,
				width=0.26\linewidth, height=5cm, scale only axis,
				group style={group size=3 by 1, horizontal sep=0.8cm,
						ylabels at=edge left},
				xmin=-1, xmax=1, xtick={\PanelXTicks},
				xticklabels={$\tfrac{1}{10}$,$\tfrac{1}{\sqrt{10}}$,$1$,$\sqrt{10}$,$10$},
				xlabel={\PanelXLabel}, ylabel={\PanelYLabel}]
			\let\origNGP\nextgroupplot%
			\renewcommand\nextgroupplot[1][]{\origNGP[#1, ymin=0.23, ymax=1.5,
					ytick={0.25,0.5,0.75,1.0,1.25,1.5}]}%
			\input{figures/PT-E-SIFT-Profile}%
			\renewcommand\nextgroupplot[1][]{\origNGP[#1, ymin=0.10, ymax=1.5,
					ytick={0.25,0.5,0.75,1.0,1.25,1.5}]}%
			\input{figures/ETH3D-F-SPLG-Profile}%
			\renewcommand\nextgroupplot[1][]{\origNGP[#1, ymin=1.2, ymax=12,
					ytick={2,4,6,8,10,12}]}%
			\input{figures/HEB-H-SIFT-Profile}%
			\let\nextgroupplot\origNGP%
		\end{groupplot}
	\end{tikzpicture}
	\par\vspace{0.4em}
	\centerline{%
		\begin{tikzpicture}[font=\footnotesize]
			\matrix [matrix of nodes, ampersand replacement=\&,
				column sep=0.3em, nodes={anchor=west, inner sep=1pt}] {
				\tikz[baseline=-0.6ex]{\draw[color=ColorRANSAC, line width=1.5pt] (0,0) -- (0.9em,0);} \& \ransac \&
				\tikz[baseline=-0.6ex]{\draw[color=ColorMSAC, line width=1.5pt] (0,0) -- (0.9em,0);} \& \msac \&
				\tikz[baseline=-0.6ex]{\draw[color=ColorGaU, line width=1.5pt] (0,0) -- (0.9em,0);} \& \gau \&
				\tikz[baseline=-0.6ex]{\draw[color=ColorMAGSAC, line width=1.5pt] (0,0) -- (0.9em,0);} \& \magsacpp \&
				\tikz[baseline=-0.6ex]{\draw[color=black, line width=1.0pt] (0,0) -- (0.9em,0);} \& Oracle \&
				\& \emph{Ours:} \&
				\tikz[baseline=-0.6ex]{\draw[color=ColorJeffreys, line width=1.5pt] (0,0) -- (0.9em,0);} \& Jeffreys \&
				\tikz[baseline=-0.6ex]{\draw[color=ColorIGRobust, line width=1.5pt] (0,0) -- (0.9em,0);} \& \igrobust \&
				\tikz[baseline=-0.6ex]{\draw[color=ColorIGMLE, line width=1.5pt] (0,0) -- (0.9em,0);} \& \igmle \\
			};
		\end{tikzpicture}%
	}
	\caption{\textbf{Profile} across three datasets, one per
		geometric problem: PT~/~$\bE$ (left), \ethd~/~$\bF$
		(middle), \heb~/~$\bH$ (right).  Each panel plots, as
		solid lines, the median rotation error across all pairs
		in the dataset versus each method's hyperparameter on a
		log axis, swept exhaustively over a two-decade interval.
		Color-matched dashed horizontal lines mark each method's
		achievable performance when its hyperparameter is free
		to be tuned per pair to select the model closest to the
		ground truth; a large gap to the solid curve indicates
		hyperparameter-dependent model switching.  Filled circles mark
		each method's minimum on the fine grid; the axis is normalized
		by the coarse-search optimum, so $x = 1$ need not be a minimum.
		Shaded bands are 95\% bootstrap confidence intervals on the median.
		\textbf{In the data-rich regime (PT, \ethd) our variants
			are remarkably robust to hyperparameter
			miscalibration}, with curves nearly flat across the full
		two-decade window where threshold-based baselines degrade
		markedly.  On \heb---the data-scarce
		regime---\textbf{the \igrobust prior regularizes the
			score and outperforms the baselines over most of the
			hyperparameter interval}
		(cf.~\Cref{par:pseudo-obs}).}
	\label{fig:profile-grid}
\end{figure}

The panels of \Cref{fig:profile-grid} plot median error against
$\kappa/\kappa^\star$, so the same horizontal decade represents
comparable parameter neighbourhoods across methods.  The bands
come from $1\,000$ pair-resamples of the dataset; because
validation and test share the pool, they measure the precision
of the fixed-pool median, not a generalization gap (that is the
Sensitivity experiment's job).  The solid black line is the
GT-Oracle, the lowest error in each pair's pool, i.e.\ the lower
bound for any scoring function.

\subsubsection{Results}
In the data-rich regime (\phototourism and \ethd), our variants
are remarkably robust to miscalibration of their
hyperparameter~$a$: the median-error curves are nearly flat
across the full two-decade window---varying by ${<}0.02^\circ$ on
\phototourism---while the threshold-based baselines degrade as
the hyperparameter is perturbed (on \phototourism they span
${\sim}0.8^\circ$--$2.9^\circ$ over the window, vanilla \ransac most
sharply).  Our three variants (Jeffreys, \igmle, \igrobust)
overlap throughout the window: \igrobust fixes $\alpha_0 = 1$ and
\igmle fits an $\alpha_0$ that is likewise small relative to the
per-pair inlier count ($n_p \approx 140$ on PT), so the data term
dominates the posterior~\eqref{eq:ig-posterior} and the
informative variants approach the Jeffreys limit
(\Cref{par:pseudo-obs}).  A held-out version of this experiment, with
each hyperparameter tuned on validation scenes and then perturbed by up
to $128\times$ on \phototourism test scenes, shows the same separation:
our score retains $\SelStuMAAInfl$ mAA@$10^\circ$ at $128\times$
inflation, against at most $\SelSimfitMAAInfl$ for the baselines
(\Cref{sec:supp-miscal}).

\paragraph{Convergence at the optimum.}
Our variants and the threshold-based baselines reach
the \emph{same} ceiling on every dataset, despite using
different inlier distributions: the marginalized model has a
Student-$t$ inlier (from the Gaussian + IG hierarchy after
integrating $\sigma^2$ analytically), while the
threshold-based baselines have a Gaussian inlier at a
globally pinned $\sigma$.  Two effects
collapse the gap as $n_p$ grows.  First, the per-partition
posterior on $\sigma^2$,
$\mathrm{IG}\!\bigl(\alpha_0 + \tfrac{\nu_p}{2},\,
	\beta_0 + \tfrac{\mathrm{RSS}_p}{2}\bigr)$, has shape
$\alpha_0 + \nu_p/2 \to \infty$, so the Student-$t$ inlier has
$2\alpha_0 + \nu_p$ degrees of freedom and converges to a
Gaussian centered at the per-partition scale estimate
$\hat s_p$.  Second, $\hat s_p$ has $\sqrt{2/\nu_p}$ relative
standard error across pairs, so it concentrates around a
population-typical value.  An optimally-tuned $\kappa$ pins
$\sigma$ globally to that same population-typical value, and
the model selected per pair coincides with what the marginal
score selects.  What no fixed $\sigma$ can reproduce is the
flatness under miscalibration above; on \heb the prior no longer
recedes and this flatness does not hold, the regime analyzed
next.

\paragraph{Low-cardinality regime.}
\heb stresses a second axis of difficulty: in addition to a
low inlier ratio (which controls \ransac's iteration count),
it has a low cardinality of inlying matches per pair, which
controls the statistical reliability of any per-partition
scale estimate.  Mechanically, its annotations are per-plane
rather than per-pair, so a typical pair carries hundreds of
geometrically valid correspondences but only those on the
annotated dominant plane (median $n_p \approx 10$, a
${\sim}4\%$ ratio) are GT inliers.  The remaining ${\sim}96\%$ are \emph{not}
mismatches; rather they are correct matches from competing
surfaces, but from the perspective of a particular dominant
plane they appear as coherently structured outliers, which
violate the uniform-outlier assumption shared by the
marginal-scoring and threshold-based families.  Imperfect assignment of spatially-verified point matches to
dominant planes injects further label noise, raising the
floor of the per-pair noise-variance estimate
$\hat\sigma^2_p$ (the variance of inlier residuals under the
oracle pose, used to anchor the IG prior).  In this context, \igrobust pulls away from the baselines and
the other variants over most of the two-decade
search region.  With $\alpha_0 = 1$ contributing two
pseudo-residuals and a median-anchored $\beta_0$, the prior
regularizes per-partition scale estimates when $n_p$ is
comparable to the model's degrees of freedom.  The median
anchor is itself insensitive to the inflated $\hat\sigma^2_p$
tails produced by structured-outlier contamination and by
annotation errors in the dominant-plane labels.  MLE regression of the IG parameters is not robust to these
corruptions, which negates \igmle's regularization advantage.
Jeffreys gives up the regularization entirely, and with only
$n_p \approx 10$ inliers per pair its
per-partition scale estimate is too noisy to discriminate
models reliably.  Threshold-based methods cannot adapt at
all, since a single global $\sigma$ has no per-partition
correction.

\paragraph{Oracle-based scoring.}
The per-pair-best envelope (dashed line) reports each method's
achievable error if its hyperparameter were retuned on every
pair to select the model closest to the ground truth.  These
envelopes invert the ordering of the solid curves' minima:
\ransac gives the best lower bound, threshold-based baselines
next, our variants highest.
This is not a measure of attainable performance but an
artifact of how each family's hyperparameter interacts with
the per-pair argmax.  For threshold-based methods the
hyperparameter is a \emph{model-selection} knob: different
thresholds induce different inlier sets and select different
models, so a per-pair sweep explores a wide range of
selections.  For our IG marginal, $a$ enters only the
outlier-penalty term $-(N - n_p)\,d_g\log(2a)$; in the data-rich
regime the \sortsweep picks an inlier partition at the
natural residual gap, largely insensitive to $a$, so sweeping
$a$ shifts the score's value but not which model is selected,
and the IG per-pair-best collapses to a single fixed model.
This decoupling is the desired property: $a$ calibrates
score values while the selected model is governed by the
data, not the hyperparameter.  The threshold-based baselines'
lower dashed envelope is therefore not a deployable
benefit---it is the visible footprint of the same dual-role
mechanism that makes their solid curves so sensitive to
hyperparameter perturbation.  In deployment one
hyperparameter is used per method, so the relevant
comparison is between the solid-curve minima; the dashed
envelope is a per-pair oracle no real scoring function has
access to.

\subsection{Sensitivity}
\label{sec:sensitivity-expt}

A \ransac user typically has only a few ground-truth examples
available for tuning and selecting their method's
hyperparameter.  Small validation sets do not fully capture
the test distribution, so one scoring method may appear
better than another by chance, depending on which validation
pairs were drawn~\cite{Shekhovtsov2025}.  We measure
\emph{expected} test performance by randomly drawing small
validation sets from the data.

We compute this expected test performance via bootstrap.  For
each image pair we precompute the rotation error of every
method at every grid point in the two-decade hyperparameter
window from \Cref{sec:eval-protocol}: $1\,000$ minimal-sample
models are drawn per pair, scored at every hyperparameter
$\kappa$, the best-scoring model is selected at each $\kappa$,
and its error against the GT pose is recorded.  With these errors
cached, the bootstrap loop is fast: draw a validation set
$\mathcal{V}$ of cardinality $|\mathcal{V}|$ (on \heb, sampled
from a random subset of scenes held out for validation, with the
disjoint remaining scenes forming the test pool, so trial
variance reflects cross-scene generalization; split at the pair
level on \phototourism and \ethd) and pick the
validation-optimal hyperparameter
$\kappa^\star(\mathcal{V}) = \arg\min_\kappa
	\mathrm{median}_{p \in \mathcal{V}}\, e_p(\kappa)$, and
evaluate $e^{\mathrm{test}}(\kappa^\star(\mathcal{V})) =
	\mathrm{median}_{p \in \mathcal{T}}\, e_p(\kappa^\star(\mathcal{V}))$
on the held-out test pool $\mathcal{T}$.  Repeating $1\,000$
times gives the expected median test error
$\mathbb{E}[e_{\mathbf{R}}]$ at each validation budget
$|\mathcal{V}| \in \{2, 4, 8, \ldots, 1024\}$ (extended to
$2048$ on \heb).

\begin{figure}[t]
	\centering
	\input{figures/sensitivity_styles_shared}%
	\input{figures/sensitivity_styles}%
	\pgfplotsset{panel_axis/.append style={
				label style={font=\scriptsize},
				title style={font=\scriptsize},
				tick label style={font=\tiny}}}%
	\pgfplotsset{ours_jeffreys_marker/.append style={mark options={solid}}}%
	\pgfplotsset{ours_ig_robust_marker/.append style={mark options={solid}}}%
	\pgfplotsset{ours_ig_mle_marker/.append style={mark options={solid}}}%
	\begin{tikzpicture}
		\begin{groupplot}[panel_axis,
				width=0.40\linewidth, height=1.9cm, scale only axis,
				group style={group size=2 by 3, horizontal sep=1.4cm,
						vertical sep=1.2cm, xlabels at=edge bottom},
				xmode=log, log basis x=2,
				xmajorgrids,
				xtick={4,16,64,256,1024},
				xlabel={Validation set size},
				scaled y ticks=false,
				yticklabel style={/pgf/number format/.cd, fixed, precision=1},
				every axis y label/.style={
						at={(axis description cs:-0.13,0.5)},
						anchor=center, rotate=90, font=\scriptsize}]
			\let\origNGP\nextgroupplot%
			\renewcommand\nextgroupplot[1][]{\origNGP[#1, ylabel={$\mathbb{E}[e_{\mathbf{R}}]$ (deg)}, ymax=1, ytick={0.8,0.9,1}]}%
			\input{figures/PT-E-SIFT-Sensitivity-eRmean}%
			\renewcommand\nextgroupplot[1][]{\origNGP[#1, ylabel={$\mathrm{std}[e_{\mathbf{R}}]$ (deg)}, ymax=0.2, ytick={0,0.1,0.2}]}%
			\input{figures/PT-E-SIFT-Sensitivity-eRstd}%
			\renewcommand\nextgroupplot[1][]{\origNGP[#1, ylabel={$\mathbb{E}[e_{\mathbf{R}}]$ (deg)}, ymax=0.6, ytick={0.2,0.4,0.5,0.6}]}%
			\input{figures/ETH3D-F-SPLG-Sensitivity-eRmean}%
			\renewcommand\nextgroupplot[1][]{\origNGP[#1, ylabel={$\mathrm{std}[e_{\mathbf{R}}]$ (deg)}, ymax=0.2, ytick={0,0.1,0.2}]}%
			\input{figures/ETH3D-F-SPLG-Sensitivity-eRstd}%
			\renewcommand\nextgroupplot[1][]{\origNGP[#1, ylabel={$\mathbb{E}[e_{\mathbf{R}}]$ (deg)}, ymax=10, ytick={7,8,9,10}]}%
			\input{figures/HEB-H-SIFT-Sensitivity-eRmean}%
			\renewcommand\nextgroupplot[1][]{\origNGP[#1, ylabel={$\mathrm{std}[e_{\mathbf{R}}]$ (deg)}, ytick={1,2,3}]}%
			\input{figures/HEB-H-SIFT-Sensitivity-eRstd}%
			\let\nextgroupplot\origNGP%
		\end{groupplot}
	\end{tikzpicture}
	\par\vspace{0.4em}
	\newcommand{\SensSwatch}[2]{%
		\tikz[baseline=-0.6ex]{%
			\draw[color=#1, line width=1.5pt] (0,0) -- (0.9em,0);%
			\draw[color=#1, line width=1.0pt] plot[mark=#2, mark size=3pt, mark options={solid}] coordinates {(0.45em,0)};%
		}%
	}%
	\noindent\hbox to \linewidth{\footnotesize
		\SensSwatch{ColorRANSAC}{x}\,\ransac\hfill
		\SensSwatch{ColorMSAC}{diamond*}\,\msac\hfill
		\SensSwatch{ColorGaU}{triangle*}\,\gau\hfill
		\SensSwatch{ColorMAGSAC}{star}\,\magsacpp\hfill
		\emph{Ours:}\hfill
		\SensSwatch{ColorJeffreys}{pentagon}\,Jeffreys\hfill
		\SensSwatch{ColorIGRobust}{o}\,\igrobust\hfill
		\SensSwatch{ColorIGMLE}{square}\,\igmle%
	}
	\caption{\textbf{Sensitivity}: expected test rotation error
		(left) and its standard deviation (right) as a function
		of validation-set size, for three cells spanning the
		geometric-problem axis---PT~/~$\bE$ (top), \ethd~/~$\bF$
		(middle), \heb~/~$\bH$ (bottom).  Shaded bands are 95\%
		BCa bootstrap confidence intervals over the
		cross-validation trials.
		\textbf{Our methods are nearly insensitive to validation-set
			size}: even with as few as 2--4 validation pairs the IG prior
		regularizes the per-partition scale enough to recover
		near-optimal hyperparameters, while threshold-based baselines
		need $\ConvRatioPT{\times}$ more validation data to converge on \phototourism
		($\ConvRatioETH{\times}$ on \ethd at a 5\% tolerance).
		On \heb---where per-pair inlier counts are too sparse to
		support a stable joint-MLE prior fit---\textbf{\igrobust's
			median-only anchor} is the appropriate variant.}
	\label{fig:sensitivity-grid}
\end{figure}

\subsubsection{Results}
Our variants reach near-optimal expected test error
at validation budgets as small as $|\mathcal{V}| = 2$, while
the threshold-based baselines need $\ConvRatioPT\times$ more
validation data to converge on \phototourism ($\ConvRatioETH\times$ on \ethd
at a 5\% tolerance).  On \phototourism the three IG
variants (Jeffreys, \igmle, \igrobust) sit within
$0.006^\circ$ of their large-$|\mathcal{V}|$ asymptote of
$0.77^\circ$ already at $|\mathcal{V}| = 2$, whereas
\msac, \gau, and \magsacpp start at ${\sim}0.90^\circ$
and need $|\mathcal{V}| \approx 1024$ to reach the same
floor; vanilla \ransac starts higher ($1.08^\circ$) and never
reaches it, plateauing near $0.87^\circ$.  \ethd shows the same convergence pattern.  Our variants are also more reproducible
at small $|\mathcal{V}|$: their cross-trial standard deviation
$\mathrm{std}[e_{\mathbf{R}}]$ sits at
$0.002^\circ$--$0.019^\circ$ at $|\mathcal{V}| = 2$, while
threshold-based baselines disperse over
$\BaselinesStdVtwoLo^\circ$--$\BaselinesStdVtwoHi^\circ$ at the same budget---every baseline at least $\StdGapVtwoLo\times$ (\ethd) to
$\StdGapVtwoHi\times$ (\phototourism) more than any proposed variant.

This is the deployment counterpart of the Profile result:
our variants' curves in \Cref{fig:profile-grid} are nearly flat,
so the $\kappa^\star(\mathcal{V})$ chosen on a tiny validation set
lands in the wide ``good $\kappa$'' region with high probability,
whereas the baselines' steeper curves leave a narrow one that a small
$\mathcal{V}$ rarely hits.  For the \ransac
practitioner this means our variants can be deployed
with a handful of ground-truth pairs---or, in the data-rich
regime, with the prior alone---while threshold-based scoring
demands a substantial labelled validation set to be tuned
reliably.

\paragraph{\heb inverts the prior-irrelevance story.}
On \heb, Jeffreys is the worst of the three variants at
small $|\mathcal{V}|$ ($9.58^\circ$ at $|\mathcal{V}| = 2$ vs.\
$8.18^\circ$ for \igrobust), with the largest trial-to-trial std
of any method ($2.84^\circ$): this is the low-$\nu_p$ regime of
\Cref{sec:profile}, where a per-partition scale estimate from
$n_p \approx 10$ inliers cannot discriminate models without prior
regularization.  \igrobust's median-anchored prior supplies it
without being dragged by the heavy $\hat\sigma^2_p$ tails that
corrupt \igmle's joint fit, and dominates \heb at every
$|\mathcal{V}|$ in mean ($7.65^\circ$ asymptote vs.\ $7.85^\circ$
for the best threshold-based baseline) and in std (lowest of any
method for $|\mathcal{V}| \ge 32$).

\paragraph{Complete pipeline.}
The selected prefix also supplies the inlier ratio and scale that
local optimization and termination need (\Cref{sec:supp-deploy-lo});
in one complete \ransac pipeline (\Cref{sec:supp-deploy-e2e}) the score is more accurate than
\superansac~\cite{Barath2025superansac} under every tested scale
miscalibration, and faster at calibration ($\EtoEOursMs$ versus
$\EtoESRMs$\,ms per pair) and under inflation of the scale.

\section{Conclusion}
\label{sec:conclusion}

We marginalize the inlier scale analytically out of the
\gaussunif mixture under a conjugate \invgamma
prior, yielding a closed-form \ransac score, computable by
\sortsweep, in which the inlier scale is genuinely absent and
a single expression spans the Jeffreys limit and informative
empirical-Bayes priors.  In the data-rich regime the score is
nearly insensitive to hyperparameter miscalibration where
threshold-based scoring degrades (\Cref{fig:profile-grid}) and
reaches near-optimal test error from two validation pairs
(validation selects each method's hyperparameter; the disjoint
test split is used only for evaluation) where the baselines need
$\ConvRatioPT\times$ more (\Cref{fig:sensitivity-grid}); in the
data-scarce regime the prior supplies regularization that is
automatically downweighted as data grows.  For the practitioner,
a handful of ground-truth pairs, or none where Jeffreys
suffices, is enough to deploy near-optimally.

\paragraph{Limitations.}\mbox{}\\
The score is inlier-scale-free, not hyperparameter-free: the outlier
half-width~$a$ remains, though it is far less sensitive than a
threshold (\cref{eq:implicit-threshold}, \Cref{sec:profile}).  The informative variants need
annotated correspondences to fit their prior; Jeffreys needs none but
is less reliable when the selected inlier set is small.  The selected
degrees of freedom~$\nu^\star$ flag this at inference, since the
posterior of~$\sigma^2$ has relative spread about
$\sqrt{2/\nu^\star}$, though a large~$\nu^\star$ does not detect
structured outliers.  The \gaussunif model
does not describe structured outliers such as \heb's competing
planes.  Our evaluation is limited to two-view geometry and mostly to fixed
candidate pools.

\begin{ack}
	This work was supported by Kiel Training for Excellence, an EU
	Horizon Europe programme under the Marie Sk\l{}odowska-Curie Actions
	(MSCA), grant agreement No~101081480, through funding of James Pritts,
	and by KMS Kiel Marine Science -- Centre for Interdisciplinary Marine
	Science at Kiel University.  The authors declare no competing
	interests.
\end{ack}

\bibliographystyle{plainnat}
\bibliography{main}

\newpage
\appendix
\setcounter{section}{0}
\renewcommand{\thesection}{\Alph{section}}

\section*{Supplementary Material}

\section{Derivation of the Profile-Marginal Score}
\label{sec:supp-score-deriv}

For a fixed partition~$\cI$, multiplying the partition
likelihood~\eqref{eq:joint-partition} by the \invgamma prior
$\pi(\sigma^2) \propto (\sigma^2)^{-(\alpha_0+1)}\exp(-\beta_0/\sigma^2)$
and integrating gives
\begin{equation}
	p(\br \mid \btheta, \cI)
	\;\propto\;
	\frac{1}{(2a)^{n_O d_g}\,(2\pi)^{n_I d_g/2}}
	\int_0^\infty
	\frac{
		\exp\!\bigl({-}\tfrac{\RSS_\cI/2 + \beta_0}{\sigma^2}\bigr)
	}{
		(\sigma^2)^{n_I d_g/2 + \alpha_0 + 1}
	}\,
	d\sigma^2,
	\label{eq:marg-sigma}
\end{equation}
where $\RSS_\cI \coloneqq \sum_{i \in \cI}\|\br_i\|^2$
is the residual sum of squares over inliers and the
proportionality absorbs the prior normalizer
$\beta_0^{\alpha_0}/\Gamma(\alpha_0)$, which is constant in
$(\btheta, \cI)$ and drops out of the partition optimization.
The integral evaluates to the \invgamma normalizing
constant
\begin{equation}
	\int_0^\infty
	(\sigma^2)^{-(n_I d_g/2 + \alpha_0 + 1)}
	\exp\!\Bigl({-}\frac{\RSS_\cI/2 + \beta_0}{\sigma^2}\Bigr)\,
	d\sigma^2
	= \Gamma\!\Bigl(\frac{n_I\,d_g}{2} + \alpha_0\Bigr)
	\Bigl(\frac{\RSS_\cI}{2} + \beta_0\Bigr)^{\!-(n_I d_g/2 + \alpha_0)}.
	\label{eq:invgamma-integral}
\end{equation}
At $(\alpha_0,\beta_0) = (0,0)$ this collapses to
$\Gamma(n_I d_g/2)\,(\RSS_\cI/2)^{-n_I d_g/2}$, the standard
Jeffreys-marginalized form.

Replacing $n_I\,d_g$ by the effective degrees of freedom
$\nu = n_I\,d_g - n_\theta$ (the $n_\theta$ parameters consumed by
the minimal-sample fit) and taking logarithms gives the boxed
closed form~\eqref{eq:score}; \Cref{sec:supp-dof}
derives this correction.

\paragraph{The variance estimate~$\hat s^2$.}
Under the posterior $\sigma^2 \mid \cI \sim \IG(\alpha, \beta)$
of~\eqref{eq:ig-posterior}, with $\alpha = \alpha_0 + \nu/2$ and
$\beta = \beta_0 + \RSS_\cI/2$, the precision $\sigma^{-2}$ is
Gamma-distributed with shape~$\alpha$ and rate~$\beta$.  Equivalently,
the same normalizing integral as in~\eqref{eq:invgamma-integral}, with
the exponent raised by one, gives
$\mathbb{E}[\sigma^{-2} \mid \cI]
	= \frac{\beta^\alpha}{\Gamma(\alpha)}\cdot
	\frac{\Gamma(\alpha+1)}{\beta^{\alpha+1}} = \alpha/\beta$, so
$\hat s^2 = \beta/\alpha = (2\beta_0 + \RSS_\cI)/(2\alpha_0 + \nu)$.
Unlike the posterior mean $\beta/(\alpha-1)$ and mode
$\beta/(\alpha+1)$, it carries no offset and so equals the mean square
of the real and pseudo-observations exactly; it is also the variance
that reproduces the posterior-averaged Gaussian exponent,
$\mathbb{E}[\RSS_\cI/2\sigma^2 \mid \cI] = \RSS_\cI/2\hat s^2$.

\subsection{The Degrees-of-Freedom Correction}
\label{sec:supp-dof}

The score~\eqref{eq:score} uses $\nu = n_I\,d_g - n_\theta$ rather
than $n_I\,d_g$, although the partition
likelihood~\eqref{eq:joint-partition} is written for a fixed
model~$\btheta$.  We now derive this correction and state the
assumptions under which~\eqref{eq:score} is the exact log of a
well-defined score.

\paragraph{Setting.}
A minimal sample~$\cS$, $|\cS| = m$, yields the candidate
$\hat\btheta_\cS = f(\bx_\cS)$.  We assume an \emph{exact minimal
solver}, $m\,d_g = n_\theta$ (true for every model in
\Cref{tab:model-dims}), so that
\begin{equation}
	\br_i(\hat\btheta_\cS) = \bzero, \qquad i \in \cS,
	\label{eq:dof-interp}
\end{equation}
with multi-root solvers handled one root at a time.  All residuals
below are evaluated at~$\hat\btheta_\cS$.  A partition splits the
observations into inliers $\cI \supseteq \cS$ and outliers~$\cO$;
$\cR = \cI \setminus \cS$ are the $n_I - m$ non-sample inliers.
Inlier residuals follow the Gaussian and outlier residuals the uniform
component of~\eqref{eq:component-densities}.

\paragraph{The score is a predictive likelihood.}
Writing $\bx_{\setminus\cS}$ for the observations outside the sample,
the score of a candidate and partition is
\begin{equation}
	\Lambda(\cS, \cI) = p(\bx_{\setminus\cS} \mid \bx_\cS, \cS, \cI).
	\label{eq:dof-predictive}
\end{equation}
Because~\eqref{eq:dof-interp} forces the candidate to interpolate its
own sample, no non-degenerate density can be assigned to~$\bx_\cS$
under its own candidate, so conditioning on the sample is forced:
$\Lambda$ is an out-of-sample predictive score, with the status of
leave-$m$-out cross-validation, not a marginal evidence.

\paragraph{Plug-in.}
Replacing the posterior over~$\btheta$ by a point mass at the solver
output, $p(\btheta \mid \bx_\cS, \sigma^2) \approx
\delta_{\hat\btheta_\cS}(\btheta)$, while keeping a prior on the noise
scale gives
\begin{equation}
	\Lambda(\cS, \cI) \approx \int_0^\infty
	p(\bx_{\setminus\cS} \mid \hat\btheta_\cS, \sigma^2, \cI)\,
	\pi(\sigma^2)\, d\sigma^2,
	\label{eq:dof-plugin}
\end{equation}
which amounts to the working assumption
$\br_i(\hat\btheta_\cS) \mid \sigma^2 \overset{\text{iid}}{\sim}
\mathcal{N}(\bzero, \sigma^2 \mathbf{I}_{d_g})$ for $i \in \cR$.

\paragraph{Conditioning out the sample.}
Given~$\bx_\cS$, both $\hat\btheta_\cS$ and, by~\eqref{eq:dof-interp},
$\br_\cS = \bzero$ are deterministic; only~$\br_\cR$ is random.  For
every bounded~$g$,
\begin{equation}
	\mathbb{E}\bigl[g(\br_\cS, \br_\cR) \mid \bx_\cS, \sigma^2, \cI\bigr]
	= \int g(\bzero, \br_\cR)\;
	p(\br_\cR \mid \hat\btheta_\cS, \sigma^2, \cI)\, d\br_\cR .
	\label{eq:dof-condition}
\end{equation}
The sample coordinates therefore enter with weight exactly one---not
with the density value $(2\pi\sigma^2)^{-m d_g/2}$, since they were
forced to zero by the fit rather than observed there---and contribute
nothing to the residual sum of squares: $\RSS_\cI = \RSS_\cR$.

\paragraph{Likelihood.}
With $n_O = N - n_I$ outliers,
\begin{equation}
	p(\br_\cR, \br_\cO \mid \hat\btheta_\cS, \sigma^2, \cI)
	= (2\pi\sigma^2)^{-\nu/2}
	\exp\!\Bigl({-}\frac{\RSS_\cI}{2\sigma^2}\Bigr)\,
	(2a)^{-n_O d_g},
	\quad
	\nu = (n_I - m)\,d_g = n_I\,d_g - n_\theta,
	\label{eq:dof-likelihood}
\end{equation}
using $m\,d_g = n_\theta$: exactly the $n_\theta$ Gaussian
coordinates consumed by the fit are removed, while the RSS is
unchanged.

\paragraph{Marginalization.}
With $\sigma^2 \sim \IG(\alpha_0, \beta_0)$, $\alpha_0, \beta_0 > 0$,
the \invgamma integral~\eqref{eq:invgamma-integral} at
$\alpha = \alpha_0 + \nu/2$, $\beta = \beta_0 + \RSS_\cI/2$ gives
\begin{equation}
	\Lambda(\cS, \cI)
	= (2\pi)^{-\nu/2}\,(2a)^{-n_O d_g}\,
	\frac{\beta_0^{\alpha_0}}{\Gamma(\alpha_0)}\,
	\frac{\Gamma(\alpha_0 + \nu/2)}
	{(\beta_0 + \RSS_\cI/2)^{\alpha_0 + \nu/2}},
	\label{eq:dof-marginal}
\end{equation}
whose logarithm is~\eqref{eq:score} up to the prior normalizer
$\beta_0^{\alpha_0}/\Gamma(\alpha_0)$, which is common to all
partitions and candidates.  The score is finite for every partition:
$\nu = 0$ gives the all-outlier baseline $(2a)^{-(N-m)d_g}$, and
$\RSS_\cI \to 0$ stays bounded because $\beta_0 > 0$.  Letting
$\alpha_0, \beta_0 \to 0$ and dropping the improper constant, again
common to all partitions so that the ranking is preserved, recovers
the Jeffreys form~\eqref{eq:score-jeffreys},
\begin{equation}
	\Lambda_J(\cS, \cI) \;\propto\;
	\frac{\Gamma(\nu/2)}{(\pi\,\RSS_\cI)^{\nu/2}}\,(2a)^{-n_O d_g},
	\qquad \nu > 0,\ \RSS_\cI > 0.
\end{equation}

\paragraph{Scope.}
The argument requires exact minimal candidates.  When $d_g$ does not
divide~$n_\theta$ (e.g.\ a homography with one distortion parameter,
$n_\theta = 9$, $d_g = 2$), the minimal sample over-determines
the model, its residuals are not all zero, and the interpolation
step~\eqref{eq:dof-interp} fails.  After local
optimization~\eqref{eq:dof-interp} no longer holds, $\RSS_\cI =
\RSS_\cR$ fails with it, and the same count $\nu = n_I\,d_g -
n_\theta$ follows instead from the classical projection argument for
a least-squares fit on~$\cI$.

\section{The Score's Implicit Threshold}
\label{sec:implicit-threshold}

We derive the implicit threshold~\eqref{eq:implicit-threshold} for
general codimension~$d_g$ and prior $(\alpha_0, \beta_0)$.  Write
$\nu_k = k\,d_g - n_\theta$, $\RSS_k = \sum_{j=1}^{k} \delta_{(j)}^2$,
$A_k = \alpha_0 + \nu_k/2$ and $B_k = \beta_0 + \RSS_k/2$ for the
prefix $\cI_k$ of \Cref{alg:score}, so that
$\hat s_k^2 = B_k/A_k$ is the variance estimate of \Cref{sec:score}.

\paragraph{Exact increment.}
Moving the next point, at distance $\delta = \delta_{(k+1)}$, from
the outliers to the inliers adds $d_g$ residual coordinates, adds
$\delta^2$ to the residual sum of squares and removes one outlier,
so by~\eqref{eq:score}
\begin{equation}
	\begin{aligned}
		\Delta S_k \coloneqq S(\btheta, \cI_{k+1}) - S(\btheta, \cI_k)
		 & = \log\frac{\Gamma(A_k + d_g/2)}{\Gamma(A_k)}
		- \Bigl(A_k + \frac{d_g}{2}\Bigr)\log\Bigl(B_k + \frac{\delta^2}{2}\Bigr) \\
		 & \quad + A_k \log B_k - \frac{d_g}{2}\log(2\pi) + d_g \log(2a).
	\end{aligned}
	\label{eq:delta-S}
\end{equation}
Since $-(A_k + \tfrac{d_g}{2})\log(B_k + \tfrac{\delta^2}{2}) + A_k\log B_k
= -\tfrac{d_g}{2}\log B_k - (A_k + \tfrac{d_g}{2})\log(1 + \tfrac{\delta^2}{2B_k})$,
the increment is exactly $\Delta S_k = \log\bigl(\pin^*(\br)/\pout\bigr)$
at $\|\br\| = \delta$, where
\begin{equation*}
	\pin^*(\br) = \frac{\Gamma(A_k + d_g/2)}{\Gamma(A_k)\,(2\pi B_k)^{d_g/2}}
	\Bigl(1 + \frac{\|\br\|^2}{2B_k}\Bigr)^{-(A_k + d_g/2)}
\end{equation*}
is the posterior predictive density of a new residual, a $d_g$-variate Student-$t$ with
$2A_k$ degrees of freedom and scale~$\hat s_k$.  This holds for every
prefix, including the small ones just beyond the minimal sample.

\paragraph{Large-sample limit.}
As $A_k$ grows the Student-$t$ tends to the Gaussian: with
$\log\Gamma(A_k + d_g/2) - \log\Gamma(A_k) \approx (d_g/2)\log A_k$
and $(A_k + d_g/2)\log(1 + \delta^2/2B_k) \approx A_k\,\delta^2/2B_k$,
\begin{equation*}
	\Delta S_k \;\approx\;
	-\frac{d_g}{2}\log\bigl(2\pi\hat s_k^2\bigr)
	- \frac{\delta^2}{2\hat s_k^2}
	+ d_g\log(2a)
	\;=\;
	\log\frac{\pin(\br;\hat s_k)}{\pout}\Big|_{\|\br\| = \delta}.
\end{equation*}
Adding a point thus raises the score exactly when the Gaussian with
the current variance estimate explains it better than the uniform
outlier density, the MAP rule of the \gaussunif mixture at
$\gamma = \tfrac12$.  The increment decreases in~$\delta$, so along
the sorted sweep it changes sign once: the optimal prefix ends where
$\pin(\br;\hat s) = \pout$ at $\|\br\| = \tau^*$, with
$\delta_{(k^*)} \le \tau^* < \delta_{(k^*+1)}$.  Solving for~$\tau^*$
gives~\eqref{eq:implicit-threshold}.  For finite~$\nu$ the
Student-$t$'s heavier tails place the exact boundary slightly further
out.

\section{Estimating the Prior Parameters \texorpdfstring{$(\alpha_0, \beta_0)$}{(alpha0, beta0)}}
\label{sec:supp-prior-estimation}

The empirical-Bayes variants fit the prior from per-pair scale
estimates $\{\hat\sigma^2_p\}_p$ on the validation set.  For pair~$p$
with labelled inlier set $\cI_p^{\mathrm{gt}}$ of size
$n_p$, we fit the model to $\cI_p^{\mathrm{gt}}$ by an
unconstrained least-squares fit (DLT for~$\bH$) and set
$\hat\sigma^2_p = \RSS_p^{\mathrm{gt}} / (d_g\,n_p - n_\theta)$, with
$\RSS_p^{\mathrm{gt}}$ the Sampson sum of squares of the inliers under that fit.
We do not fit under the oracle pose (for~$\bH$, only the 3-DOF plane
normal): \heb's labelled inliers are not strictly coplanar, so that
estimate inflates by orders of magnitude on a non-trivial fraction of
pairs.  For relative pose the two fits coincide.  Pairs with $d_g\,n_p \le n_\theta$ yield no
$\hat\sigma^2_p$ and are excluded from the prior aggregation; they
remain test pairs, since the score~\eqref{eq:score} only requires the
selected partition to satisfy $\nu > 0$.

\paragraph{\igmle: type-II maximum likelihood.}  Using the identity
$\sigma^2 \!\sim\! \IG(\alpha,\beta) \iff
	1/\sigma^2 \!\sim\! \mathrm{Gamma}(\alpha,\,1/\beta)$,
we fit a Gamma distribution to $\{1/\hat\sigma^2_p\}_p$ and
read off $(\alpha_0, \beta_0)$ from its shape and inverse
scale.  Both parameters are jointly identified, so the prior
captures both the location and the dispersion of the
validation $\hat\sigma^2_p$.  This estimator is preferred when
the per-pair $\hat\sigma^2_p$ are well concentrated, but is
sensitive to outliers: a pair with few inliers gives a
$\hat\sigma^2_p$ with relative standard error $\sqrt{2/\nu_p^{\mathrm{gt}}}$,
$\nu_p^{\mathrm{gt}} = d_g\,n_p - n_\theta$, and a small number of extreme values can drag
the joint MLE far from the bulk of the distribution.

\paragraph{\igrobust: median anchor.}  We fix $\alpha_0 = 1$, i.e.\
two pseudo-observations, and set
$\beta_0 = \alpha_0 \cdot \mathrm{median}(\hat\sigma^2_p)$, so that the
prior mean square $s_0^2 = \beta_0/\alpha_0$ is the median of
$\{\hat\sigma^2_p\}$.  This sacrifices the
dispersion information captured by joint MLE in exchange for
robustness to heavy tails in $\{\hat\sigma^2_p\}$, which arise
on datasets with widely-varying per-pair inlier counts.

Both estimators are recomputed once per cross-validation trial
and held fixed for both the $a$-selection sweep and the test
evaluation.

\section{Relation to Torr's Bayesian Framework, \gric, and \magsacpp}
\label{sec:supp-torr}

\paragraph{Torr's Bayesian manifold formulation.}
Torr's Bayesian treatment of epipolar geometry and generic manifold
fitting~\cite{Torr2002} already shows that marginalizing the latent
noiseless correspondence removes the tangential error, leaving a
Gaussian likelihood in the manifold codimension~$d_g$, together with a
uniform outlier model.  This is the foundation our
likelihood~\eqref{eq:component-densities} builds on.

\paragraph{\gric.}
\gric addresses \emph{model-class} selection: it approximates the
evidence for manifolds of different dimensions and parameter counts
using a fixed-scale \mapsac/\msac data term plus complexity penalties
for the latent point coordinates and the global model parameters.  In
our setting the model class is fixed, so these penalties are constant
across hypotheses, while the data term still requires an externally
specified~$\sigma$.  Our contribution is complementary: starting from
the same codimension-based \gaussunif likelihood, we condition
on the candidate and its minimal sample, profile over the inlier
partition, and integrate~$\sigma^2$ exactly under an \invgamma
prior.  \gric approximates model-class evidence; our score removes the
inlier scale for within-class \ransac scoring.  Our $\nu = n_I\,d_g -
n_\theta$ counts residual Gaussian evidence after fitting
(\Cref{sec:supp-dof}); it is not \gric's complexity penalty.

\paragraph{The \magsacpp likelihood.}
\magsacpp~\cite{Barath2020} begins from a $\chi_4$ density because a
two-view correspondence has four coordinates.  The residual, however,
is a distance to the model manifold, so only the $d_g$ orthogonal
constraint directions remain ($d_g = 1$ for essential and fundamental
matrices, $d_g = 2$ for homographies): the relevant dimension is the
model codimension, not the ambient observation dimension
(\Cref{sec:supp-dims}).  More fundamentally, even $\chi_{d_g}$ is not
an appropriate model-ranking likelihood.  It is the density of the
\emph{radius} of a Gaussian residual vector and includes the volume
factor $r^{d_g-1}$; because $r = r(\btheta)$ depends on the candidate,
this is a parameter-dependent transformation of the observations, and
for $d_g \geq 2$ it assigns zero density to an exact fit.  Our
formulation instead starts from the Gaussian density of the
$d_g$-dimensional residual vector itself: $d_g$ is the per-observation
constraint dimension and $\nu$ the total residual evidence after
fitting, and neither is used as the shape of a radial likelihood.

		\section{Model Dimensions: Parameters, Codimension, and Minimal Sample Size}
		\label{sec:supp-dims}

\Cref{tab:model-dims} lists the quantities of \Cref{sec:partition}:
$n_\theta$ is the number of free parameters after fixing gauge
freedoms, $d_g$ the codimension, $m = \lceil n_\theta/d_g\rceil$ the
minimal sample size (the Jacobian
$\pdv{\bg_\cI}!{\btheta} \in \Reals^{k\,d_g \times n_\theta}$ generically
has rank~$n_\theta$ only for $k \ge m$), and $d$ the dimension of an
observation~$\bx_i$.

\begin{table}[ht]
	\centering
	\caption{Dimensional quantities for standard two-view models.
		$n_\theta$: model parameters (free DOF);
		$d_g$: codimension (constraints per point);
		$m = \lceil n_\theta/d_g\rceil$: minimal sample size;
		$d$: observation dimension.}
	\label{tab:model-dims}
	\smallskip
	\begin{tabular}{lcccc}
		\toprule
		\textbf{Model}                    & $n_\theta$ & $d_g$ & $m$ & $d$ \\
		\midrule
		Fundamental matrix ($\mathbf{F}$) & 7          & 1     & 7   & 4   \\
		Homography ($\mathbf{H}$)         & 8          & 2     & 4   & 4   \\
		Essential matrix ($\mathbf{E}$)   & 5          & 1     & 5   & 4   \\
		\bottomrule
	\end{tabular}
\end{table}

\section{Scoring Runtime}
\label{sec:supp-deploy}
\label{sec:supp-deploy-runtime}

\Cref{tab:runtime-hyp} reports single-threaded per-hypothesis
timings on \phototourism.  All scorers receive identical precomputed residuals.  For
\magsacpp we use the lookup-table implementation prescribed by the
paper and shipped with the reference code, its most optimistic
faithful runtime; the closed form is included for completeness.
Although comparison sorting is $O(N\log N)$, our implementation sorts
with a radix sort on 32-bit float keys, so the \sortsweep is
$\Theta(N)$: increasing the correspondence count by $2.21\times$
increases the runtime by $2.24\times$, against the $2.48\times$
that $N\log N$ scaling would predict.  Our score costs $8.8$--$19.7\,\mu$s more per hypothesis than \msac
and $7.8$--$16.6\,\mu$s more than the \magsacpp lookup table.

\begin{table}[ht]
	\centering
	\caption{Per-hypothesis scoring time (single thread, identical
		precomputed residuals) at two mean correspondence counts~$\bar n$.}
	\label{tab:runtime-hyp}
	\smallskip
	\begin{tabular}{llrr}
		\toprule
		Method   & Form                 & $\bar n = 715$     & $\bar n = 1580$    \\
		\midrule
		\ransac   & exact                & $0.65\,\mu$s       & $1.43\,\mu$s       \\
		\msac     & exact                & $0.84\,\mu$s       & $1.88\,\mu$s       \\
		\magsacpp & lookup table         & $1.83\,\mu$s       & $5.00\,\mu$s       \\
		Ours     & exact \sortsweep & $9.62\,\mu$s       & $21.59\,\mu$s      \\
		\gau      & exact                & $12.38\,\mu$s      & $28.49\,\mu$s      \\
		\magsacpp & closed form          & $29.73\,\mu$s      & $92.99\,\mu$s      \\
		\bottomrule
	\end{tabular}
\end{table}

These microbenchmarks are not end-to-end slowdowns.  A complete
\ransac iteration also includes minimal solving, residual evaluation
and local optimization, and the termination criterion determines how
many times all of these costs are paid.  \Cref{sec:supp-deploy-e2e}
therefore measures a complete pipeline.

\section{Scale Miscalibration}
\label{sec:supp-miscal}

The main experiments isolate the scoring function on fixed candidate
pools with each method's hyperparameter tuned on the data. This section
measures how the scores (\Cref{sec:supp-scoring}), and a complete \ransac
pipeline built on ours (\Cref{sec:supp-deploy-e2e}), degrade when the
user-supplied scale is wrong.

\paragraph{Scale corruption.}
Methods are tuned once at calibration. At test time, each user-supplied
quantity expressed in residual units is multiplied by
$\zeta\in[1/128,128]$ (deflation for $\zeta<1$, inflation for $\zeta>1$);
dimensionless multipliers and scales recomputed from the selected partition
are unchanged. Thus $\zeta$ measures sensitivity to external calibration
rather than perturbing every internal variable identically.

\paragraph{Metric.}
Unlike the main experiments, which report the orientation error
(\Cref{sec:eval-protocol}), this section reports the pose error, which also
includes the translation direction. For relative pose, we evaluate the four
essential-matrix decompositions and report the smallest maximum angular
error~\cite{Jin2021}. We define
mAA@$10^\circ$ as
\begin{equation}
	\begin{aligned}
		\mathrm{mAA}@10^\circ
		 & =\frac{1}{10^\circ}\int_0^{10^\circ}
		P(e_{\mathrm{pose}}\le t)\,dt,                                 \\
		\text{s.t.}\quad e_{\mathrm{pose}}
		 & =\min_{j\in\{1,2,3,4\}}\max\bigl(e_R^{(j)},e_t^{(j)}\bigr).
	\end{aligned}
	\label{eq:maa}
\end{equation}
Here $e_R^{(j)}$ and $e_t^{(j)}$ are the rotation and
translation-direction errors of decomposition~$j$. In~\eqref{eq:maa},
$t$ is the pose-error threshold, and $P(e_{\mathrm{pose}}\le t)$ is the
per-scene fraction of pairs whose pose error is at most~$t$. We compute
mAA per scene and then average over scenes, with failed estimates
contributing zero.

\paragraph{Parameter mapping.}
\label{sec:supp-scale-mapping}

The shared scoring coordinate $q\in[1,3.3]$ and the validation reference
scale~$\hat\sigma$, the square root of the median per-pair noise
variance~$\hat\sigma_p^2$ (\Cref{sec:supp-prior-estimation}) over the
validation pairs, determine each score's native residual-unit parameter;
$q$ plays the role of the quantile in $\tau = q\sigma$ (\Cref{sec:intro}).
\ransac, \msac, \simfitpp, and \gau use a threshold or cutoff
$\tau=q\hat\sigma$. \magsacpp uses the upper noise-scale bound
$\bar\sigma=q\hat\sigma$, not a residual cutoff. For the profile-marginal
score, the outlier half-width is
\begin{equation}
	a(q;\hat\sigma,d_g)=\hat\sigma\sqrt{\frac{\pi}{2}}
	\exp\!\left(\frac{q^2}{2d_g}\right),
	\label{eq:native-scale-maps}
\end{equation}
which equates the isotropic Gaussian inlier density and uniform outlier
density at residual norm $q\hat\sigma$ under equal assignment priors, the
inverse of the implicit threshold~\eqref{eq:implicit-threshold}. This is a
calibration map, not the exact boundary of the profile-marginal partition;
generatively, $a$ remains an outlier-domain parameter.

Replacing $\hat\sigma$ by $\zeta\hat\sigma$ at fixed $q^*$ multiplies each
native parameter by~$\zeta$:
\begin{equation}
	\begin{aligned}
		\tau(q^*;\zeta\hat\sigma)          & =\zeta\tau(q^*;\hat\sigma),             \\
		\bar\sigma(q^*;\zeta\hat\sigma)    & =\zeta\bar\sigma(q^*;\hat\sigma),       \\
		a(q^*;\zeta\hat\sigma,d_g)         & =\zeta a(q^*;\hat\sigma,d_g).
	\end{aligned}
	\label{eq:native-scale-corruption}
\end{equation}
The scoring-only experiment (\Cref{sec:supp-scoring}) applies the nearest
evaluation-grid value to every test pair without retuning. Inverting the
maps at the original reference scale~$\hat\sigma$ gives the equivalent
coordinates
\begin{equation}
	q_{\tau}^{\prime}=q_{\bar\sigma}^{\prime}=\zeta q^*,
	\qquad
	(q_a^{\prime})^2=(q^*)^2+2d_g\log\zeta,
	\label{eq:corrupted-cutoff}
\end{equation}
where $q_a^{\prime}$ is real only when the right-hand side is nonnegative.

\subsection{Scoring}
\label{sec:supp-scoring}

This experiment evaluates how well each score selects a model under scale
miscalibration, without local optimization: each score ranks the identical
pool of minimal-sample hypotheses of each image pair, and the selected
hypothesis is evaluated directly. Both experiments of this section use a
separate set of \phototourism pools: the 21 scenes are split by scene, not by
pair as in \Cref{sec:experiments}, into \PTValScenes{} validation and
\PTTestScenes{} test scenes of $2\,200$ pairs each, with \rootsift
correspondences capped at $4\,096$ and $4\,000$ minimal-sample hypotheses per
pair. Ours is the
profile-marginal score with the Jeffreys prior. Cross-validation uses only
the shared scoring window $q\in[1,3.3]$, a dimensionless multiple of the
validation reference scale~$\hat\sigma$ (\Cref{sec:supp-scale-mapping}). At
test time the supplied scale is multiplied by $\zeta\in[1/128,128]$ with the
tuned coordinate~$q^*$ held fixed; each score's parameter follows through
its map (\Cref{sec:supp-scale-mapping}) and is used for all pairs without
retuning. For the profile-marginal score, the affected parameter is the
outlier half-width~$a$; the selected partition and its variance
estimate~$\hat s^2$ remain data-dependent. \simfitpp~\cite{Edstedt2025} is
\msac with a data-driven threshold: starting from an initial threshold, it
alternates fitting on a random half of the correspondences with a
bias-corrected median scale estimate on the other half and scores at the
converged threshold; its initial threshold is cross-validated on the~$\tau$
axis like the other threshold-based scores, and the estimate falls back to
it when the iteration leaves the admissible range. Because that range is the
reference's $0.25$--$8$\,px window, \simfitpp scores the pool's residuals
re-expressed in pixels through the stored image normalizations; every other
score works in the normalized frame, where all of its parameters are
defined. The shaded confidence intervals in \cref{fig:supp-sel-miscal} use a
scene-clustered BCa bootstrap~\cite{Efron1987}.

\begin{figure}[ht]
	\centering
	\input{figures/sel_miscal_colors}%
	\begin{subfigure}[t]{0.48\linewidth}
		\centering
		\caption{median pose error}
		\input{figures/panel_sel_miscal}
	\end{subfigure}\hfill
	\begin{subfigure}[t]{0.48\linewidth}
		\centering
		\caption{mAA@$10^\circ$}
		\input{figures/panel_sel_miscal_maa}
	\end{subfigure}\\[3pt]
	\pgfplotslegendfromname{selmiscallegend}
	\caption{\textbf{Scoring under scale miscalibration, without local
			optimization} (\phototourism essential matrices; \PTTestScenes{}
		test and \PTValScenes{} validation scenes). We report (a) the
		scene-averaged median pose error, with the axis clipped at $6^\circ$
		so that the near-tied calibrated regime stays readable (the
		threshold-based scores leave the panel under inflation and reach
		$\SelThreshMedInflLo^\circ$--$\SelThreshMedInflHi^\circ$ at
		$128\times$), and (b) mAA@$10^\circ$~\eqref{eq:maa} at the same
		operating points, which are tuned on the median. The baselines are
		\ransac~\cite{Fischler1981}, \msac~\cite{Torr1998},
		\magsacpp~\cite{Barath2020}, \gau~\cite{Shekhovtsov2025}, and
		\simfitpp~\cite{Edstedt2025}, which re-estimates the inlier threshold
		per pair from held-out residuals of the hypothesis selected on a
		random half of the correspondences.}
	\label{fig:supp-sel-miscal}
\end{figure}

At calibration ($\zeta=1$) the scores are nearly tied
(\cref{fig:supp-sel-miscal}): the profile-marginal score and \simfitpp reach
median errors of $\SelStuMedCal^\circ$ and $\SelSimfitMedCal^\circ$ and mAA
of $\SelStuMAACal$ and $\SelSimfitMAACal$. At $128\times$ inflation, the
profile-marginal score reaches a median error of $\SelStuMedInfl^\circ$ and
$\SelStuMAAInfl$ mAA, while \simfitpp reaches $\SelSimfitMedInfl^\circ$ and
$\SelSimfitMAAInfl$ mAA; the other threshold-based scores have median
errors of $\SelThreshMedInflLo^\circ$--$\SelThreshMedInflHi^\circ$ and at
most $\SelThreshMAAInfl$ mAA. The profile-marginal score is thus
substantially more stable under scale corruption. This reflects both its
data-dependent partition selection and scale estimation; by
\eqref{eq:corrupted-cutoff}, the equivalent coordinate of the half-width also
changes only with~$\log\zeta$, $(q_a')^2 = (q^*)^2 + 2d_g\log\zeta$, and the
experiment does not isolate these effects. Re-estimating the threshold per pair (\simfitpp) does not
remove the dependence on the supplied scale: at $128\times$ inflation its
median error of $\SelSimfitMedInfl^\circ$ is close to that of the other
threshold-based scores, against $\SelStuMedInfl^\circ$ for the
profile-marginal score.

These pools hold more hypotheses than the scores need: the budget of 1000
minimal samples per pair used in \Cref{sec:experiments} suffices for every
scorer: on 8\,800 \phototourism pairs with about
4\,000 \rootsift correspondences each, from four held-out scenes, all scorers tie at correct calibration
(mAA@$10^\circ$ 0.534--0.542, median error $2.80^\circ$--$2.96^\circ$),
and under $32\times$ miscalibration ours retains 0.466 mAA and
$4.60^\circ$ while the baselines fall to 0.10--0.14 mAA and
$20^\circ$--$25^\circ$.

\subsection{End-to-End \ransac}
\label{sec:supp-deploy-e2e}

The deployed pipeline (Ours) combines the profile-marginal score with
local optimization whose loss width is anchored to the selected partition
(\Cref{sec:supp-deploy-lo}). We compare it with
\superansac~\cite{Barath2025superansac}, including sampling and adaptive
termination, on the \PTTestScenes{} \phototourism test scenes
(\Cref{tab:e2e}). Both are tuned at calibration; at test time, $\zeta$
multiplies each system's residual-unit parameter. At calibration, ours
reduces the median pose error from $\EtoESRMed^\circ$ to
$\EtoEOursMed^\circ$ and raises mAA@$10^\circ$ from $\EtoESRMAA$ to
$\EtoEOursMAA$. Under corruption, our median error stays within
$\EtoEOursRange^\circ$, whereas \superansac's reaches $\EtoESRRangeHi^\circ$
under deflation and $\EtoESRInflMed^\circ$ at $128\times$ inflation. We
postulate that the calibrated gain has the same origin: a threshold tuned
on validation pairs is effectively miscalibrated for test pairs whose noise
scale lies outside the validation distribution, whereas partition anchoring
estimates the scale per pair.

\begin{table}[ht]
	\centering
	\small
	\setlength{\tabcolsep}{4pt}
	\caption{\textbf{RANSAC under scale miscalibration.}
		Scene-averaged median pose error and mAA@$10^\circ$ on the
		\PTTestScenes{} \phototourism test scenes; median wall time per pair,
		single-threaded, on a stratified sample of \EtoETimePairs{} pairs per
		scene (median of \EtoETimeRepeats{} repeats); both systems tuned at
		$\zeta=1$, \superansac at its validated maximum noise level with at
		most $10\,000$ trials. Bold marks the better system per column.}
	\label{tab:e2e}
	\begin{tabular}{@{}llrrrrr@{}}
 \toprule
 & $\zeta$ & $1/16$ & $1$ & $4$ & $16$ & $128$ \\
 \midrule
 Ours & error [$^\circ$] & \textbf{2.69} & \textbf{1.57} & \textbf{1.84} & \textbf{2.59} & \textbf{6.20} \\
 & mAA@$10^\circ$ & \textbf{0.595} & \textbf{0.686} & \textbf{0.667} & \textbf{0.620} & \textbf{0.503} \\
  & time [ms] & 291 & \textbf{50} & \textbf{34} & \textbf{23} & \textbf{15} \\
 \midrule
 \superansac~\cite{Barath2025superansac} & error [$^\circ$] & 53.73 & 2.23 & 3.49 & 10.14 & 38.27 \\
  & mAA@$10^\circ$ & 0.210 & 0.643 & 0.558 & 0.333 & 0.072 \\
  & time [ms] & \textbf{86} & 111 & 168 & 265 & 381 \\
 \bottomrule
\end{tabular}

\end{table}

\section{Local Optimization and Termination}
\label{sec:supp-deploy-lo}

The \sortsweep supplies every quantity the downstream components
of \ransac need: the selected support~$\cI^\star$, its cutoff
$\tau^\star = \delta_{(k^\star)}$, the inlier ratio $\hat w = k^\star/N$,
and the variance estimate~$\hat s^2$ of \Cref{sec:score}.
Standard confidence-based termination stops after
\begin{equation}
	K = \frac{\log(1-p)}{\log(1-\hat w^{\,m})}
\end{equation}
minimal samples, for a confidence~$p$ that an all-inlier sample has been
drawn, and local optimization uses either~$\tau^\star$ or a
partition-anchored width $\lambda\hat s$, with only the
dimensionless~$\lambda$ cross-validated.  No calibrated residual
threshold is required at test time.

\Cref{tab:lo-anchor} isolates this effect with the same \poselib
truncated-least-squares Levenberg--Marquardt optimizer (TLS-LM)~\cite{PoseLib},
on the \phototourism pools of \Cref{sec:supp-miscal}.  The fixed
variant uses a calibrated threshold~$\tau_{\mathrm{ref}}$, perturbed
at test time by~$\zeta$; the anchored variant re-estimates
$\tau = \lambda\hat s$ for each hypothesis.  Both are tuned at
$\zeta = 1$.

\begin{table}[ht]
	\centering
	\caption{Local optimization with a fixed versus a partition-anchored
		width.  Median pose error [deg] over the 11 \phototourism test scenes; $\zeta$
		multiplies the calibrated residual unit at test time.}
	\label{tab:lo-anchor}
	\smallskip
	\begin{tabular}{lrrr}
		\toprule
		\poselib TLS-LM                                & $\zeta = 1$ & $\zeta = 16$ & $\zeta = 128$ \\
		\midrule
		fixed ($\tau_{\mathrm{ref}}$)                 & 1.8         & 4.0          & 22.0          \\
		anchored ($\tau = \lambda\hat s$)          & 1.9         & 2.4          & 4.2           \\
		\bottomrule
	\end{tabular}
\end{table}

\section{Dataset Details}
\label{sec:supp-datasets}

\Cref{tab:datasets} summarizes the three datasets, which span
three axes of variation---dataset, geometric problem, and
feature pipeline.

\begin{table}[ht]
	\centering
	\small
	\caption{Datasets used in our evaluation: geometric problem,
		feature pipeline, bootstrap protocol, number of scenes, and
		total image-pair count.}
	\label{tab:datasets}
	\begin{tabular}{l@{~}r c l@{~}r l c c}
		\toprule
		\multicolumn{2}{l}{Dataset} & Problem            & \multicolumn{2}{l}{Features} & Protocol  & Scenes                             & Pairs                          \\
		\midrule
		\phototourism               & \cite{Jin2021}     & $\bE$                        & \rootsift & \cite{Arandjelovic2012}            & Flat (resample) & 21 & 44\,211 \\
		\ethd                       & \cite{Schoeps2017} & $\bF$                        & SP+LG     & \cite{DeTone2018,Lindenberger2023} & Flat (resample) & 13 & 3\,039  \\
		\heb                        & \cite{Barath2023}  & $\bH$                        & SIFT      & \cite{Lowe2004}                    & Hierarchical    & 10 & 21\,999 \\
		\bottomrule
	\end{tabular}
\end{table}

\paragraph{Annotations.}
Each dataset provides image pairs with ground-truth absolute
pose and camera calibration, from which we construct the
ground-truth relative pose.  \heb additionally annotates the
homographies induced by the dominant planes in each scene.
\phototourism has no canonical split; we use the publicly available
scenes with ground-truth poses and SfM matches.

Each scoring method ranks a set of \emph{tentative
	correspondences} per pair; \igmle and \igrobust additionally
consume \emph{ground-truth point matches} during validation
(\Cref{sec:eval-protocol}).  \heb and \phototourism ship both a
tentative correspondence set---SIFT~\cite{Lowe2004} on \heb,
\rootsift~\cite{Arandjelovic2012} on \phototourism---and
SfM bundle-adjusted GT matches (for \heb, restricted to the
annotated plane).  These GT matches are
pseudo-ground-truth, a multi-view SfM reference treated as
gold-standard for the two-view sub-problem; they also give the
per-pair variance estimates~$\hat\sigma^2_p$ of
\Cref{sec:supp-prior-estimation}.  \ethd ships only per-view laser depth (training
subset only; the test set is unreleased); we run
\superpoint~\cite{DeTone2018}\,+\,\lightglue~\cite{Lindenberger2023}
(SP+LG), a learned detect-and-match pipeline, as the
tentative set, and verify each candidate by 3D consistency:
pixels are unprojected via the depth, brought into a common
frame by the GT poses, and accepted when the resulting 3D
points lie within a dataset-specific threshold (reflecting
feature, depth-sensor, and absolute-pose accuracy).  Pairing
SP+LG with SIFT/\rootsift on the other two datasets exercises
our variants against feature noise from both hand-designed and
learned pipelines.  On \heb a pair's matches can span several
planes, so a scoring method may select a model that fits one plane
while the GT inliers belong to another; this surfaces in some \heb
panels of the experiments.

\end{document}